\documentclass{article}

\PassOptionsToPackage{numbers,sort&compress,square}{natbib}

\usepackage[eandd,preprint]{neurips_2026}

\usepackage[utf8]{inputenc}
\usepackage[T1]{fontenc}
\usepackage{hyperref}
\usepackage{url}
\usepackage{booktabs}
\usepackage{tabularx}
\usepackage{amsfonts}
\usepackage{nicefrac}
\usepackage{microtype}
\usepackage[table]{xcolor}
\usepackage{graphicx}
\usepackage{placeins}
\usepackage{float}

\newcommand{\aucCell}[2]{\cellcolor{green!#1}#2}

\title{iMINDBench: iEEG Multi-Institution Neural Decoding Benchmark}

\author{%
  \begin{tabular}{c}
  Geeling Chau$^{1}$\thanks{Correspondence: \texttt{gchau@caltech.edu}} \quad Saba Hashemi$^{2}$\thanks{Equal contribution.} \quad Yonghyeon Gwon$^{3}$\footnotemark[2] \quad Eshani Patel$^{1,4}$ \\
  Jan DeWitt$^{5}$ \quad Christopher Wang$^{5}$ \quad Andrii Zahorodnii$^{5}$ \quad Sabera J. Talukder$^{1}$ \\
  Danny Dongyeop Han$^{3}$ \quad Chun Kee Chung$^{3}$ \quad Maryam M. Shanechi$^{2}$ \quad Yisong Yue$^{1}$ \\[0.8em]
  $^{1}$Caltech \quad $^{2}$USC \quad $^{3}$Seoul National University \\
  $^{4}$CMU \quad $^{5}$MIT
  \end{tabular}
}

\begin{document}

\maketitle

\begin{abstract}
Intracranial electroencephalography (iEEG) is widely used to record electrical activity directly from electrodes inside the human brain, making it an attractive modality for neural decoding.
However, progress in iEEG decoding, especially toward general-purpose foundation models, remains difficult to measure reliably: datasets are task- or institution-specific, limiting evidence of generalization across tasks and recording environments, and preprocessing choices can strongly influence performance, making model improvements difficult to distinguish from preprocessing gains.
Thus, we introduce \textsc{iMINDBench}, an \underline{i}EEG \underline{M}ulti-\underline{I}nstitution \underline{N}eural \underline{D}ecoding \underline{Bench}mark that evaluates models on a shared suite of fifteen decoding tasks across three naturalistic movie-watching datasets.
The benchmark additionally defines standardized preprocessing tracks and fixed evaluation splits to support consistent model comparisons.
Using iMINDBench, we find that the evaluated pretrained systems generally outperform baselines within their respective preprocessing tracks, while strong spectral baselines remain competitive across institutional datasets.
In our scaling study, adding up to 25 times more supervised data from other subjects or institutions yields only small or task-dependent gains over within-session training.
Together, these findings highlight the need for iEEG models that improve on strong preprocessing baselines and make more effective use of data across subjects and institutions.
Project website: \url{https://imindbench.github.io/}
\end{abstract}

\section{Introduction}
Neural decoding seeks models that generalize across subjects, institutions, and recording conditions while requiring minimal patient-specific calibration.
Intracranial electroencephalography (iEEG) is an attractive setting for this goal because it records neural activity directly from electrodes inside the human brain with high temporal resolution.
% %
% These properties also distinguish iEEG from scalp EEG in ways that prevent direct reuse of existing EEG benchmarks (see Related Work).
% %
However, progress in iEEG decoding remains difficult to evaluate because existing studies often differ not only in dataset and task selection, but also in preprocessing pipelines and split construction.
As a result, apparent benchmark improvements can reflect changes in evaluation setup rather than advances in neural modeling itself.

Existing resources address important parts of this problem.
Multi-institution iEEG benchmarks such as Omni-iEEG~\cite{duan2026omniieeg} and Mayo/FNUSA~\cite{nejedly2020multicenter} demonstrate that clinical recordings can be harmonized across sites, but primarily focus on seizure, pathology, high-frequency oscillation, or related epileptology tasks.
Naturalistic decoding resources such as Neuroprobe~\cite{zahorodnii2025neuroprobe}, Brain Treebank~\cite{wang2024braintreebank}, and AJILE12~\cite{peterson2022ajile12} provide interpretable naturalistic decoding tasks, but evaluations often remain tied to a single dataset or institutional setting.
Current models also differ in neural input representation and preprocessing pipeline, making improvements difficult to separate from changes in evaluation setup.

We therefore introduce iMINDBench, an \underline{i}EEG \underline{M}ulti-\underline{I}nstitution \underline{N}eural \underline{D}ecoding \underline{B}enchmark spanning fifteen language, auditory, and visual decoding tasks across naturalistic movie-watching datasets from three institutions.
The benchmark uses fixed evaluation splits and specified preprocessing tracks, with evaluations spanning scaling domains and unit decodability subsets.
Results are reported separately by dataset and preprocessing track to reveal how model performance varies across institutions and input representations.

The benchmark reveals why these controls matter.
First, the evaluated pretrained systems generally outperform non-pretrained baselines within their respective preprocessing tracks on Main units.
However, strong Multi-STFT baselines remain competitive with pretrained systems across institutional datasets.
These baselines provide a reference point for developing learned representations that match or exceed the benefits of engineered spectral features across institutions.
Second, by aligning tasks and neural-signal metadata across institutions, the benchmark lets us examine whether models benefit from additional subjects and institutions on the same task.
In our scaling study, adding other subjects from the same dataset is broadly beneficial, although the gains remain smaller than those from additional target-session data. Adding data from other datasets and institutions produces weaker and more task-dependent improvements.
These results highlight the difficulty of turning broader supervision into reliable improvements in target-session decoding.

By supporting these two complementary assessments, iMINDBench enables more systematic evaluation of general-purpose iEEG models across tasks, institutions, and preprocessing tracks, helping identify where their benefits generalize and where they remain specific to the evaluation setting.

\paragraph{Contributions.}
\begin{itemize}
  \item \textbf{Benchmark artifact.} We introduce a multi-institution iEEG decoding benchmark spanning three independently collected movie-watching datasets with fifteen aligned decoding tasks.
  \item \textbf{Evaluation contract.} iMINDBench defines fixed splits, decodability subsets, standardized Multi-STFT and Waveform preprocessing tracks, and supervised training scaling domains.
  \item \textbf{Benchmark findings.} Strong spectral baselines remain competitive with pretrained systems, providing a reference point for learned representations. Broader supervised fine-tuning yields limited, task-dependent gains in our scaling study, motivating models that learn more effectively from heterogeneous neural data.
\end{itemize}

\section{Related Work}
\label{relatedwork}
\paragraph{Datasets and benchmarks.}
Intracranial electroencephalography is attractive for naturalistic neural decoding because it records human neural activity at high temporal resolution during behavior, but clinical recordings are sparse, nonuniform, and institution-specific~\citep{parvizi2018promises}.
Existing iEEG benchmarks cover important parts of this landscape.
The Mayo/FNUSA graphoelement dataset~\cite{nejedly2020multicenter} and Omni-iEEG~\cite{duan2026omniieeg} show that multi-institution clinical iEEG can be harmonized, but focus on seizure, pathology, high-frequency oscillation, or related epileptology targets.
Naturalistic resources such as Brain Treebank~\cite{wang2024braintreebank}, Neuroprobe~\cite{zahorodnii2025neuroprobe}, BYD~\cite{keles2024multimodal}, Pippi~\cite{berezutskaya2022open}, and AJILE12~\cite{peterson2022ajile12} provide movie, language, audiovisual, motor, or activity labels, but do not jointly control task, institution, and preprocessing heterogeneity in one evaluation protocol.

While scalp EEG benchmarks such as MOABB~\cite{jayaram2018moabb} and recent EEG foundation-model evaluations~\citep{kuruppu2025eegreview,yang2026eegworth} exist, they do not transfer cleanly to iEEG: scalp EEG is volume-conducted and bandlimited, whereas iEEG records directly from cortex and retains the high-frequency activity (roughly 70--250 Hz) which is important for cognitive decoding~\citep{parvizi2018promises,lachaux2012high}. In addition, iEEG electrode placement is clinically determined rather than fixed, breaking the standardized-montage assumption behind most EEG pipelines and foundation models. These differences necessitate separate benchmarking frameworks for intracranial EEG.

\paragraph{Neural foundation models and preprocessing.}
Recent electrophysiology foundation models make shared evaluation increasingly necessary.
iEEG models such as BrainBERT~\cite{wang2023brainbert}, Brant~\cite{zhang2023brant}, PopT~\cite{chau_wang_2025_population}, BaRISTA~\cite{oganesian2025barista}, MVPFormer~\cite{carzaniga2026mvpformer}, and DIVER-1~\cite{han2026diver1}, together with EEG foundation models such as LaBraM~\cite{jiang2024labram} and EEGPT~\cite{wang2024eegpt}, address scale, subject variability, channel mismatch, spatial layout, and low signal-to-noise ratio.
However, their evaluation conditions differ across waveform inputs, spectrograms or superlets, learned tokens, frozen embeddings, spatial metadata, pretraining corpora, and split units.
A recent EEG foundation-model review argues that evaluations remain heterogeneous and limited~\citep{kuruppu2025eegreview}, and a broad EEG-FM benchmark finds that fine-tuned foundation models can perform well in data-rich settings but often do not clearly outperform compact neural or classical decoders in data-scarce regimes~\citep{yang2026eegworth}.
This benchmark targets the corresponding iEEG problem: model comparisons should be made within explicit task, institution, split, and preprocessing specifications, with matched baseline reruns whenever the input representation changes.

\section{Benchmark Design}

\begin{figure}[!htbp]
  \centering
  \includegraphics[width=\linewidth]{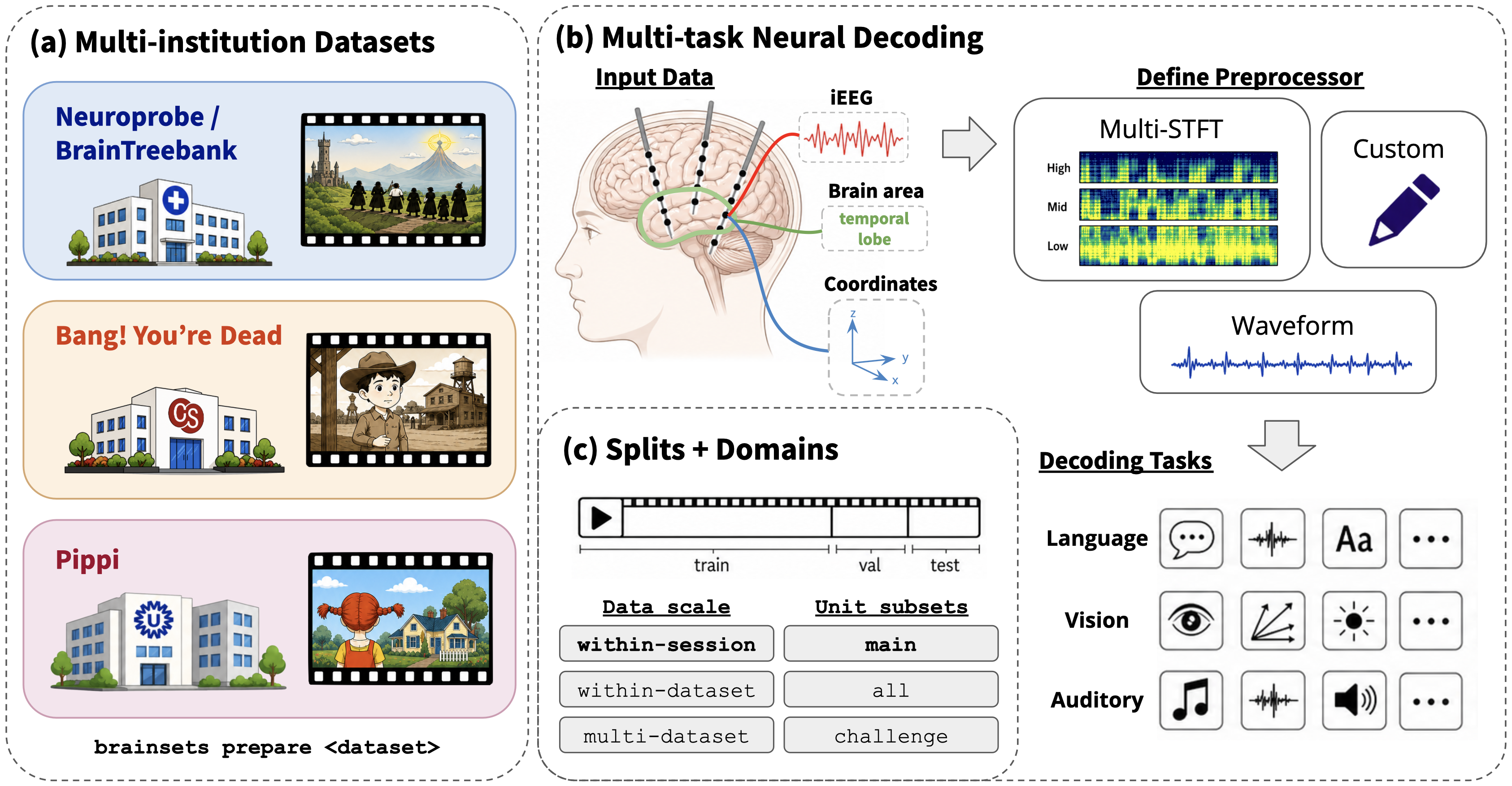}
  \caption{\textbf{Multi-institution, multi-task neural decoding benchmark.}
The benchmark aligns naturalistic decoding tasks and patient metadata across three institutions and defines preprocessing tracks, splits, and decoding domains.
(a) Three naturalistic movie-watching intracranial electroencephalography (iEEG) datasets are included: Neuroprobe, Bang! You're Dead (BYD), and Pippi.
(b) iEEG neural recordings are routed through explicit preprocessing tracks, which may include multi-resolution short-time Fourier transform (Multi-STFT), waveform, and custom preprocessors to be used as inputs to models that decode language, vision, and auditory task families.
(c) The benchmark defines challenging temporal splits, data-scale regimes, and unit decodability subsets.}
  \label{fig:overview}
\end{figure}

The benchmark is designed to evaluate whether decoding improvements survive changes in institution, preprocessing track, and behavioral target under shared evaluation settings.
To support this, iMINDBench standardizes preprocessing and evaluation across multiple naturalistic iEEG datasets while exposing performance differences across tasks and institutions.
It also supports controlled scaling experiments that test whether additional supervised data improves decoding across sessions, subjects, and institutions.
Figure~\ref{fig:overview} summarizes the benchmark workflow, including datasets, preprocessing tracks, splits, and decoding domains.

\subsection{Datasets and Tasks}

\begin{figure}[!b]
  \centering
  \includegraphics[width=\linewidth]{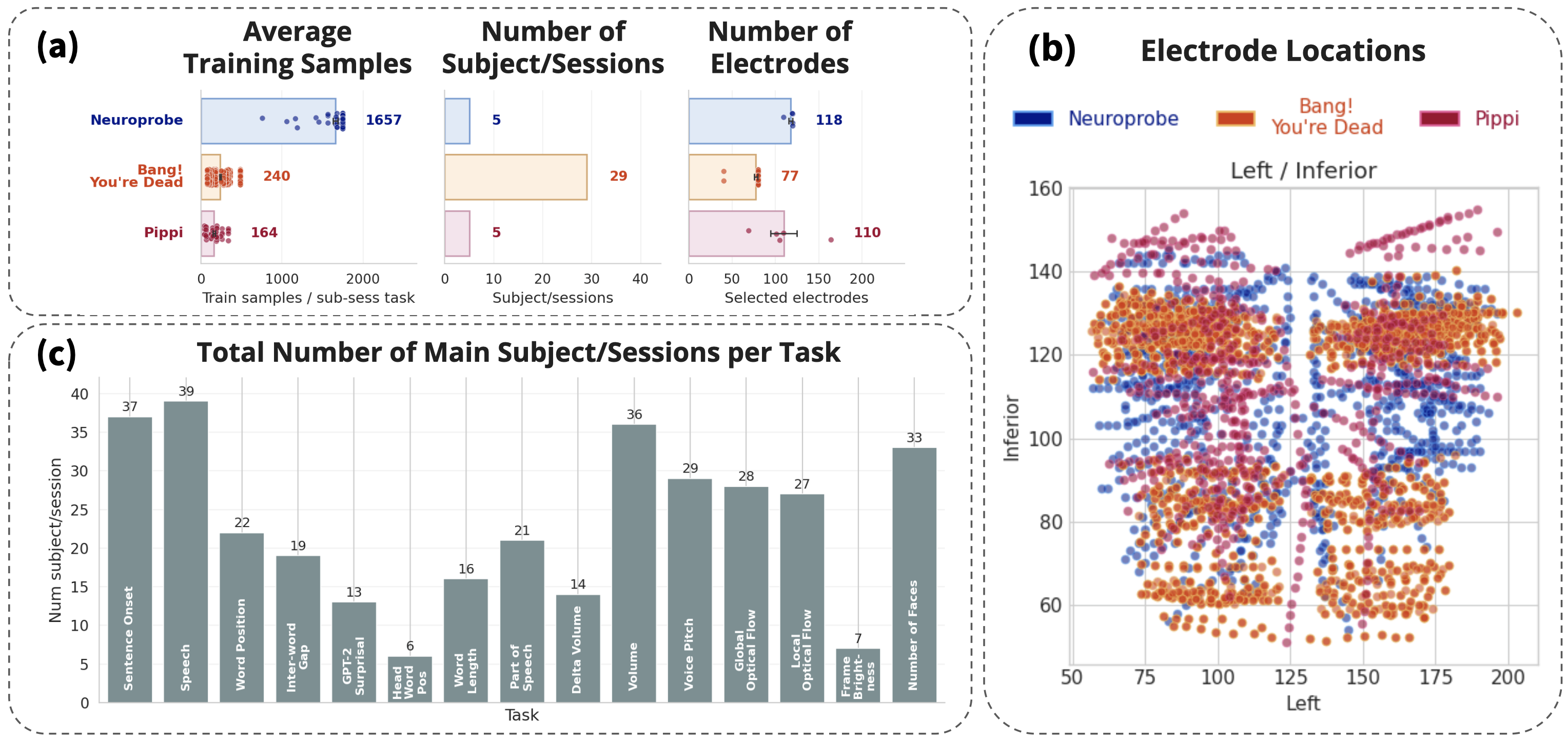}
    \caption{\textbf{Dataset coverage is heterogeneous across institutions and tasks.}
(a) Average number of training samples, Main subject/session--task units, and electrodes per subject.
(b) Electrode locations are shown in left/inferior coordinate space for the same datasets.
(c) Task bars report the number of Main subject/session--task units for each aligned decoding task.}
  \label{fig:dataset-info}
\end{figure}

\paragraph{Datasets.}
The current release combines Neuroprobe/Brain Treebank~\cite{zahorodnii2025neuroprobe,wang2024braintreebank}, the Bang! You're Dead (BYD) movie-watching dataset~\cite{keles2024multimodal}, and Pippi~\cite{berezutskaya2022open}.
Compared with the original Neuroprobe benchmark, we restrict evaluation of the Neuroprobe/Brain Treebank component to five held-out subject/session units so that remaining Brain Treebank recordings can be used for model pretraining without overlapping benchmark targets.
Each dataset contains neural recordings acquired during movie or audiovisual stimulus presentation, but the datasets differ in recording duration, electrode coverage, subject/session structure, and institutional practices.
Neuroprobe/Brain Treebank provides longer movie-watching recordings and comparatively large within-session training sets, BYD introduces a different institutional and stimulus context with shorter recordings and more explicit subject/session structure, and Pippi provides a smaller independent dataset with different electrode coverage and stimulus statistics.
Figure~\ref{fig:dataset-info} summarizes the resulting variation in dataset scale, electrode coverage, and task support across institutions.

\paragraph{Harmonization.}
iMINDBench aligns decoding targets and electrode metadata to support comparisons across institutions.
BYD and Pippi use the Neuroprobe/Brain Treebank stimulus-feature extraction pipeline, with dataset-specific annotation and timing alignment.
Available electrode coordinates and anatomical labels are harmonized to shared conventions.
Differences in acquisition, electrode layout, and coverage remain part of the evaluation setting.
Appendix~\ref{app:dataset-harmonization} describes the alignment procedures and documents the source datasets' acquisition metadata.

\paragraph{Tasks.}

The benchmark includes fifteen decoding tasks following the original Neuroprobe task definitions~\cite{zahorodnii2025neuroprobe}, spanning language, auditory, and visual domains.
These include speech presence, sentence onset, GPT-2 surprisal, pitch, volume, optical flow, brightness, and face-count decoding.
Both binary and multiclass label modes are supported; this paper reports binary classification results.
Each example pairs a 1-second neural window with a stimulus label.
Continuous annotations are binarized by contrasting low- and high-value ranges within each session, while categorical annotations are mapped to binary labels.
Complete task definitions and dataset-specific label-construction details are provided in Appendix~\ref{app:decoding-tasks}.

\subsection{Splits and Domains}
\label{sec:splits-domains}

\paragraph{Splits.}
We follow Neuroprobe's two-fold within-session splitting procedure for each dataset--subject/session--task combination, termed an evaluation \textbf{\underline{unit}}.
Applying this procedure across datasets maintains compatibility with Neuroprobe, including reuse of compatible runs for Neuroprobe submission.
Within each fold, training, validation, and test examples come from largely contiguous portions of the movie.
Preprocessing statistics are estimated from the training split and applied unchanged to validation and test data.
Each unit's final score is averaged across the two folds.

\paragraph{Scaling domains.}
Scaling domains vary the source of supervised fine-tuning data while preserving the same target units and evaluation splits.
\textbf{\underline{Within-session}} uses only target-session data, representing adaptation to a new subject/session using its calibration labels.
\textbf{\underline{Within-dataset}} adds supervised examples from other subject/session units in the same dataset.
\textbf{\underline{Multi-dataset}} further adds supervised examples from other institutions.
These domains test whether broader supervision improves decoding beyond target-session calibration.

\paragraph{Unit decodability subsets.}
{
Many evaluation units remain near chance under current baselines.
This may reflect weak task-relevant signal at the recorded electrode locations, poor recording quality, or limitations of current models in extracting task-relevant information.
We therefore report Main, Challenge, and All unit decodability subsets to distinguish improvements on units that are decodable by baseline within-session models from progress on those that are presently near chance.

We operationalize near-chance performance using two screening decoders: an STFT decoder and HTNet at 500 Hz.
The STFT decoder uses logistic regression with the single-STFT configuration that achieves the highest mean validation ROC-AUC for each unit.
A supported unit $u$ is assigned to \textbf{\underline{Main}} when
\[
\max\!\left\{
\overline{\mathrm{AUC}}_{\mathrm{val}}^{\mathrm{STFT}}(u),
\overline{\mathrm{AUC}}_{\mathrm{val}}^{\mathrm{HTNet\!-\!500\,Hz}}(u)
\right\} > 0.60,
\]
where each score is averaged across the two validation folds.
\textbf{\underline{Challenge}} contains the remaining supported units, classified as near chance because both screening scores are at or below 0.60.
\textbf{\underline{All}} contains both subsets.
These labels describe decodability under the screening models; they do not establish that Challenge units lack task-relevant neural signal.}
The resulting Main/Challenge performance separation, task-level Main-unit model breakouts, and Main subject/session coverage are summarized in Appendix Figures~\ref{fig:appendix-main-challenge-unit-performance}, \ref{fig:appendix-task-breakout-main}, and~\ref{fig:appendix-main-sub-sess}.

\subsection{Preprocessing Tracks}
\label{sec:preprocessing-tracks}

\paragraph{Track definitions.}
Decoding performance depends on both the model and how neural recordings are preprocessed, making it necessary to account for both when interpreting improvements.
Each preprocessing track specifies how neural recordings are transformed into model inputs, helping control for preprocessing variation.
We provide two complementary tracks.
The Multi-STFT track incorporates established signal-processing operations and spectral features \cite{buzsaki2012origin} that enable strong performance with simple baseline models.
The Waveform track instead provides inputs close to the raw recorded waveforms, preserving phase and temporal structure that the Multi-STFT representation does not explicitly represent.
It tests whether models can learn representations that match or exceed the benefits of engineered spectral features.
Comparisons under matched preprocessing assess model improvements using the same inputs, while comparisons across tracks assess complete decoding systems, including their preprocessing.
The benchmark also supports custom preprocessing routes, allowing compatible reference baselines to be rerun on new inputs for matched comparisons.

\paragraph{Multi-STFT track.}
The Multi-STFT track is the benchmark's standardized spectral preprocessing route.
The spectral track first applies 60 Hz + harmonics notch filtering for line noise removal and Laplacian rereferencing for common electrode noise removal as done in many prior works \cite{wang2023brainbert, li2018optimal}.
Because neural activity spans multiple temporal and frequency scales, a single fixed STFT resolution can unevenly represent different frequency bands and tasks.
The Multi-STFT representation instead uses separate STFT resolutions for low-, mid-, and high-frequency ranges before combining them into a spectrogram-like representation compatible with common neural decoding models (Appendix Figure~\ref{fig:appendix-multistft}).
The Multi-STFT sections use duration-defined windows and an approximately 62-ms hop consistently across dataset sampling rates; Appendix Figure~\ref{fig:appendix-multistft} shows the frequency ranges, and Table~\ref{tab:multistft-settings} gives the exact sample settings and retained upper bins.
The Multi-STFT track selects frequency bins within 2--250 Hz, excluding the lowest-frequency components while retaining bands including high gamma \cite{lachaux2012high}.
For each fold, the Multi-STFT track estimates normalization statistics exclusively from training samples, separately for each channel and frequency bin, and applies those fixed statistics to validation and test samples. This preserves cross-temporal information while standardizing the signals for machine-learning models.
We compare normalization strategies and their downstream performances on Multi-STFT baselines in Figure ~\ref{fig:preprocessing}a.

\paragraph{Waveform track.}
The Waveform track uses a 0.5 Hz high-pass filter, notch filtering for line noise, and Laplacian rereferencing, while permitting model-specific sampling rates and input scaling and normalization conventions. Supervised baselines use a standardized 500 Hz recipe with 60 Hz and harmonic notch filtering and robust global normalization fitted exclusively on the training split. This normalization uses a single estimated median center and a scaled median absolute deviation across training channels and timepoints, applied unchanged to validation and test samples, preserving relative amplitude differences across windows and channels.
We use 500 Hz for the supervised waveform baselines because matched comparisons did not show consistent gains at higher sampling rates. Model-specific sampling, filtering, and normalization variations are documented in Table~\ref{tab:waveform-recipes}.
Within-track comparisons involving model-specific variants evaluate complete decoding systems; they do not isolate architecture or pretraining effects.
We visualize how preprocessing changes affect baseline performance on our strongest supervised waveform model, HTNet, in Figure~\ref{fig:preprocessing}a.

\paragraph{Custom preprocessing routes.}
Users can evaluate custom preprocessing routes by supplying preprocessing code and configuration files.
These routes must be documented explicitly and, where compatible inputs or embeddings are available, evaluated with a simple matched baseline.
Leaderboard entries report preprocessing details to make these comparisons transparent (Appendix Figure~\ref{fig:leaderboard}).
Appendix~\ref{app:brainbert-scorecards} illustrates this approach for BrainBERT, comparing its pretrained representations with simple decoders using the corresponding single-STFT inputs.

\subsection{Models and Reporting}

\paragraph{Model families.}
We evaluate generic decoder baselines alongside pretrained and specialized iEEG models.
Logistic regression, multilayer perceptron (MLP), and convolutional neural network (CNN) baselines are evaluated on both Multi-STFT and Waveform inputs.
The Multi-STFT suite additionally includes pretrained PopT-v2~\cite{chau_wang_2025_population,patel2025timescale}.
The Waveform suite includes HTNet~\cite{peterson2021htnet}, trained from scratch, and pretrained BaRISTA~\cite{oganesian2025barista} and DIVER-1~\cite{han2026diver1}.
We also examined MVPFormer and Brant, whose inclusion requires further validation of adaptations from their released temporal input representations to one-second decoding windows (Appendix~\ref{app:additional-model-coverage}).
Appendix~\ref{app:model-details} provides model and training details.

\paragraph{Reporting.}
Results are reported with their preprocessing track, model family, and dataset-level performance.
For pretrained models, we also report the datasets used for pretraining to contextualize their training exposure.
For each dataset and model, we report the mean ROC-AUC across the available evaluation units.
The Overall score is the unweighted mean of the three dataset-level scores.
The benchmark website and leaderboard break down model comparisons by dataset, task, preprocessing track, scaling domain, and unit decodability subset, helping identify where model advantages are consistent and where they depend on the evaluation setting.

\section{Benchmark Results}

\subsection{Decoding Performance}

\begin{table}[!htbp]
  \centering
  \normalsize
  \setlength{\tabcolsep}{4pt}
  \renewcommand{\arraystretch}{0.9}
  \begin{tabular}{@{}lllcccc@{}}
    \toprule
    \textbf{Track} & Model & Pretraining & Overall & Neuroprobe & BYD & Pippi \\
    \midrule
    \multicolumn{7}{l}{\textbf{Multi-STFT}} \\
     & Logistic & not pretrained & \aucCell{28}{0.628} & \aucCell{15}{0.697} & \aucCell{40}{0.624} & \aucCell{40}{0.564} \\
     & MLP & not pretrained & \aucCell{33}{\underline{0.631}} & \aucCell{23}{0.702} & \aucCell{42}{\underline{0.625}} & \aucCell{43}{\underline{0.567}} \\
     & CNN & not pretrained & \aucCell{15}{0.620} & \aucCell{35}{\underline{0.710}} & \aucCell{15}{0.611} & \aucCell{15}{0.539} \\
     & PopT-v2 & BrainTreeBank & \aucCell{55}{\textbf{0.645}} & \aucCell{55}{\textbf{0.723}} & \aucCell{55}{\textbf{0.632}} & \aucCell{55}{\textbf{0.579}} \\
    \midrule
    \multicolumn{7}{l}{\textbf{Waveform}} \\
     & Logistic & not pretrained & \aucCell{15}{0.554} & \aucCell{15}{0.617} & \aucCell{15}{0.519} & \aucCell{15}{0.525} \\
     & MLP & not pretrained & \aucCell{24}{0.573} & \aucCell{28}{0.653} & \aucCell{20}{0.532} & \aucCell{23}{0.534} \\
     & CNN & not pretrained & \aucCell{26}{0.576} & \aucCell{33}{0.666} & \aucCell{21}{0.537} & \aucCell{16}{0.526} \\
     & HTNet & not pretrained & \aucCell{44}{0.614} & \aucCell{42}{0.693} & \aucCell{43}{0.597} & \aucCell{40}{0.552} \\
     & BaRISTA & BrainTreeBank & \aucCell{55}{\textbf{0.636}} & \aucCell{51}{\underline{0.716}} & \aucCell{55}{\textbf{0.632}} & \aucCell{45}{\underline{0.558}} \\
     & DIVER-1 & private+ AJILE12 & \aucCell{54}{\underline{0.634}} & \aucCell{55}{\textbf{0.728}} & \aucCell{45}{\underline{0.605}} & \aucCell{55}{\textbf{0.569}} \\
    \bottomrule
\end{tabular}
\caption{\textbf{Main benchmark results.}
Within-session ROC-AUC performance on Main units, grouped by preprocessing track.
Cells are shaded with a normalized color scheme within column and track.
Best per column and track are bolded, and second-best values are underlined.}
  \label{tab:scoreboard}
\end{table}

Table~\ref{tab:scoreboard} compares within-session decoding performance on Main units across the three datasets.
The evaluated pretrained systems outperform non-pretrained baselines within their respective preprocessing tracks.
PopT-v2 achieves the highest Overall ROC-AUC of 0.645, compared with 0.631 for the strongest non-pretrained Multi-STFT baseline.
Within the Waveform track, BaRISTA and DIVER-1 achieve 0.636 and 0.634, respectively, compared with 0.614 for HTNet.
However, the Multi-STFT MLP remains competitive with both pretrained waveform systems, particularly on BYD and Pippi, which have smaller downstream training sets than Neuroprobe.
This highlights the importance of evaluating pretrained systems against strong preprocessing baselines under limited supervision.
Because these comparisons involve different input representations, they assess complete decoding systems rather than isolating the contribution of the model.

The benefit of more complex spectral decoders varies across datasets.
On Neuroprobe, CNN outperforms both logistic regression and MLP, whereas on BYD and Pippi, logistic regression and MLP outperform CNN.
This pattern suggests that engineered spectral features support effective decoding with relatively simple models when labeled data are limited, highlighting the need for learned feature extractors that generalize reliably under limited supervision.

Waveform inputs show a different pattern.
HTNet consistently outperforms generic logistic, MLP, and CNN baselines across all three datasets, indicating that the choice of architecture matters for extracting useful information from waveform inputs.
Pretrained BaRISTA and DIVER-1 further outperform these within-session-trained baselines, although their relative strengths vary across datasets: DIVER-1 performs better on Neuroprobe and Pippi, while BaRISTA performs better on BYD.

Together, these results show that model advantages depend on the input representation and institutional dataset, motivating evaluation across both dimensions.
Additional scorecards for All and Challenge units are provided in Appendix~\ref{app:subject-subset-scorecards}.

\subsection{Preprocessing}

\begin{figure}[!t]
  \centering

  \includegraphics[width=\linewidth]{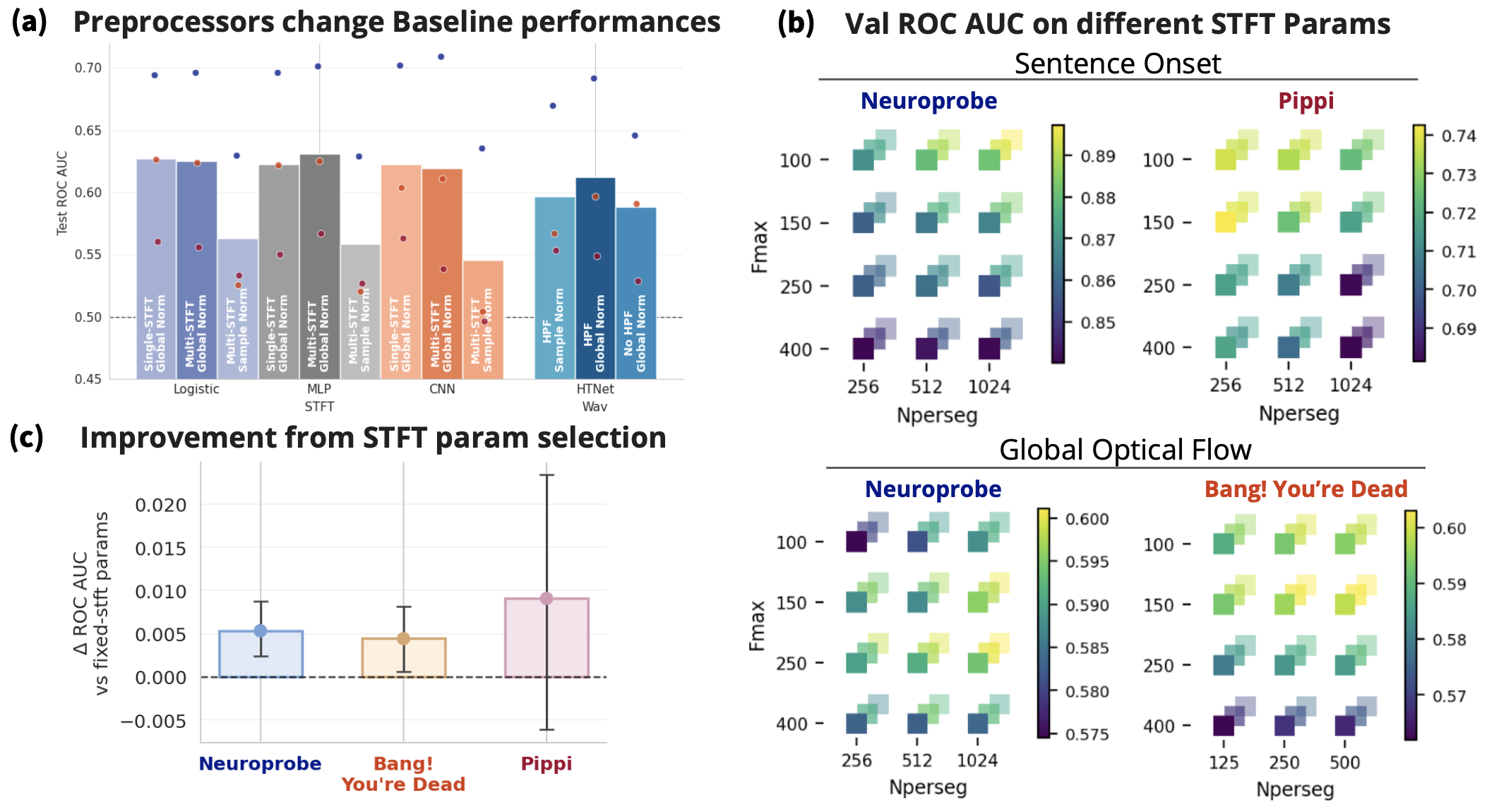}
\caption{\textbf{Preprocessing choices have large effects on baseline performance.}
(a) Baseline model performance under different STFT and Waveform preprocessing choices.
Bars show the unweighted mean of dataset-level averages; dots show individual dataset averages.
Our proposed STFT and Wav preprocessing tracks are highlighted with the darkest color in each model family.
(b) Validation ROC-AUC for STFT parameter sweeps.
We sweep max frequency (y-axis), number of timepoints per segment (x-axis) and percent overlap (depth) at 50\%, 75\%, and 87.5\%.
Different datasets (columns) may have different optimal single-STFT params for the same task (rows).
(c) Mean paired test ROC-AUC change for logistic regression using validation-selected versus fixed single-STFT parameters (150 Hz maximum frequency, 250 ms segments, 75\% overlap). Error bars indicate 95\% bootstrap confidence intervals.
All panels report results on the Main unit decodability subset.
}
  \label{fig:preprocessing}
\end{figure}

Figure~\ref{fig:preprocessing} evaluates how preprocessing choices affect benchmark conclusions under otherwise matched evaluation settings.
Baseline performance varies across preprocessing routes, indicating that neural decoding comparisons can depend strongly on input representation and normalization choices.

In Figure~\ref{fig:preprocessing}a, we compare two normalization strategies that encode different assumptions about signal scale. \emph{Sample Norm}, following approaches used in wav2vec 2.0 and BrainBERT~\cite{baevski2020wav2vec,wang2023brainbert}, standardizes each input window independently.
This may provide invariance to nuisance variation in recording gain and overall amplitude, but it can also remove between-window amplitude differences that are informative for decoding.
\emph{Global Norm} instead uses statistics estimated from the training set, preserving relative amplitude variation across examples while applying the same transformation to validation and test samples. Global Norm yields consistently stronger baseline performance, improving ROC-AUC by more than 0.10 in some evaluated settings.
This result suggests that information removed by sample-wise normalization is often task-relevant in our evaluation settings.
We therefore select Global Norm for the standardized baseline recipes in our Multi-STFT and Waveform preprocessing tracks;
Section~\ref{sec:preprocessing-tracks} details its implementation for each track.

We next test the Waveform track's sensitivity to high-pass filtering while holding the remaining HTNet preprocessing fixed.
High-pass filtering suppresses slow baseline drift and other low-frequency variation that can dominate waveform amplitude, although it may also attenuate slowly varying task-related signals.
Omitting the 0.5-Hz high-pass filter reduces mean ROC-AUC from 0.612 to 0.589, with lower performance in all three datasets (Figure~\ref{fig:preprocessing}a).
This consistent reduction suggests that, under our evaluation protocol, the benefit of suppressing very-low-frequency variation outweighs any useful information attenuated by the filter.
We therefore retain the 0.5-Hz high-pass filter in the standardized Waveform baseline recipe.

Finally, we test the spectral track's sensitivity to replacing Multi-STFT with a single STFT resolution. Aggregate performance is comparable (Figure~\ref{fig:preprocessing}a), but validation-selected Single-STFT parameters vary across datasets and tasks (Figure~\ref{fig:preprocessing}b,c). We therefore retain Multi-STFT as the flagship spectral track: its band-specific window lengths favor frequency resolution at lower frequencies and temporal resolution at higher frequencies without selecting a task- or dataset-specific resolution.

These results show that preprocessing choices can substantially affect decoding performance and support our standardized tracks through both empirical comparisons and signal-processing considerations.
Comparing models under matched preprocessing helps distinguish model improvements from gains due to preprocessing changes.

\subsection{Data Scaling}

\begin{figure}[!htbp]
  \centering
  \includegraphics[width=\linewidth]{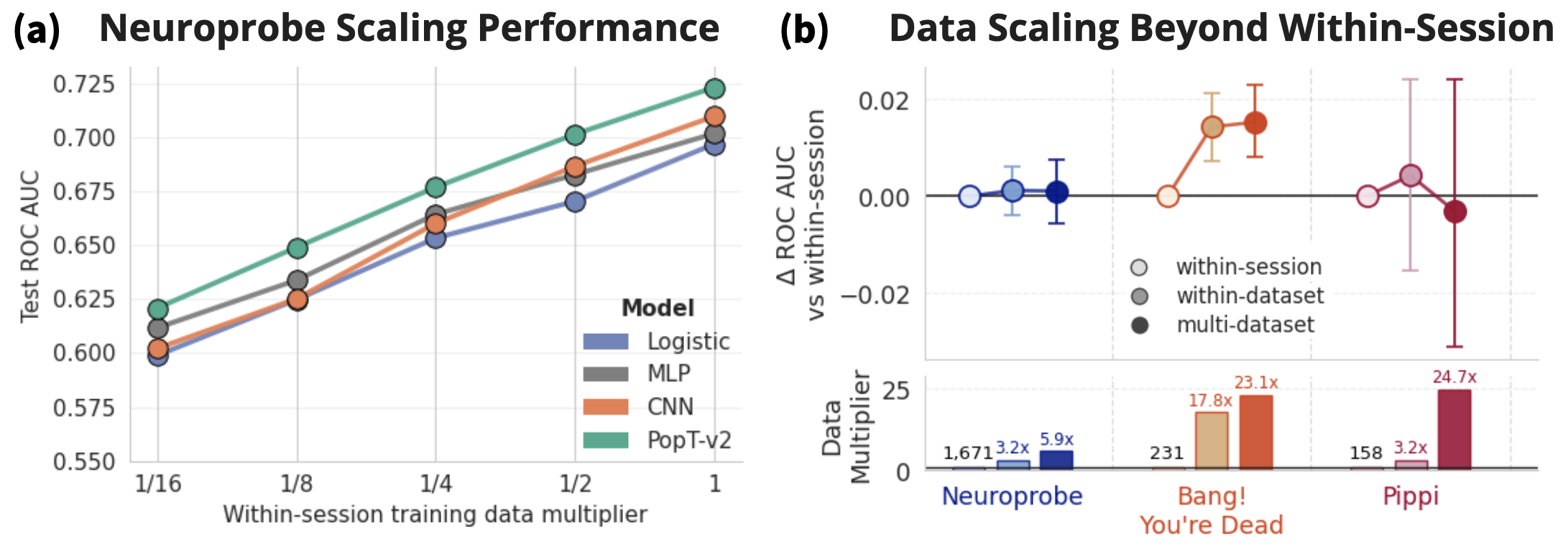}
  \caption{\textbf{Data scaling highlights a boundary in current model generalization.}
(a) Neuroprobe within-session decoding performance (y-axis) as the available training data increase (x-axis) across models evaluated on the Multi-STFT track (legend). (b) For each dataset (columns and colors), the top subplot shows the paired change in test ROC-AUC relative to within-session training (y-axis) when PopT-v2 is provided supervised data from within the same dataset (medium-shaded circles) or from multiple datasets (dark-shaded circles). The bottom subplot shows the corresponding training-data multipliers relative to within-session training (colored labels); the black label above the leftmost bar gives the mean number of within-session training samples per task. Error bars indicate 95\% confidence intervals for paired changes in ROC-AUC. Task-level decompositions are provided in Appendix Figures~\ref{fig:appendix-task-sample-efficiency} and~\ref{fig:appendix-task-scaling}. All panels report results on the Main unit decodability subset.}
  \label{fig:data-scaling}
\end{figure}

\paragraph{Transfer question.}
Data scaling is the benchmark's direct test of the transfer promise behind population and foundation-style iEEG modeling.
If broader supervised neural data are useful for naturalistic decoding, they should improve a target unit when target-session labels are scarce.
Figure~\ref{fig:data-scaling} separates that question into three data sources.
Within-session scaling measures the value of additional examples from the same target unit.
Within-dataset scaling asks whether other Main units from the same dataset help once institution, acquisition practice, stimulus family, and label construction are mostly shared.
Multi-dataset scaling is the harder test: it asks whether supervised data from other institutions and stimulus contexts improves the same target units.
We evaluate within-session scaling across spectral models and use PopT-v2, which supports joint supervised training across subjects, for within-dataset and multi-dataset scaling.

\paragraph{Scaling pattern.}
The scaling analysis exposes an important asymmetry between target-session and broader supervised data.
Across evaluated models, within-session scaling produces the largest and most consistent gains, with mean test ROC-AUC increasing by approximately 0.025 for each doubling of the training data (Figure~\ref{fig:data-scaling}a). Using approximately 3--18 times as many training samples as within-session training, within-dataset scaling improves mean ROC-AUC by 0.001 on Neuroprobe, 0.014 on Bang! You're Dead, and 0.004 on Pippi (Figure~\ref{fig:data-scaling}b). Multi-dataset training uses approximately 6--25 times as many samples as within-session training, yet underperforms within-dataset training on Neuroprobe and Pippi. On Bang! You're Dead, multi-dataset training improves mean ROC-AUC over within-dataset training by only 0.001 despite using approximately 30\% more training samples.

Within-dataset training increases mean test ROC-AUC over within-session training in 30 of 44 task--dataset pairs; adding other datasets yields further increases in 19 of 44 pairs (Appendix Figure~\ref{fig:appendix-task-scaling}). The benefit of broader supervision therefore depends on the decoding task. More effective cross-subject and cross-institution learning remains an important direction for improving downstream neural decoding.

\section{Release and Extensibility}

The evaluation code, preprocessing configurations, and baseline implementations will be released in the
\href{https://github.com/imindbench/iMINDBench}{iMINDBench repository}.
Dataset preparation pipelines and benchmark split metadata will be available through
\href{https://github.com/neuro-galaxy/torch_brain}{TorchBrain}
and its Brainsets integration, built on the Neuro-Galaxy ecosystem~\citep{azabou2023unified}.
These pipelines provide a shared interface for preparing recordings, task labels, and evaluation splits.

A publicly hosted \href{https://imindbench.github.io/}{benchmark website and leaderboard}
will support comparisons across datasets, tasks, preprocessing tracks, scaling domains, and unit decodability subsets.
To submit a model, users will run the evaluation pipeline and submit generated result artifacts and model metadata through a pull request to the leaderboard repository, where automated checks validate the submission.
Detailed instructions will be provided in the
\href{https://imindbench.github.io/leaderboard/submit.html}{submission guide}.

Additional model-native systems can be evaluated through the custom preprocessing route when coverage is complete and matched baseline reruns are available.
Future releases may add datasets, tasks, preprocessors, models, or hidden-test evaluation while preserving archived benchmark versions.

\section{Discussion}

The results show that preprocessing choices and cross-dataset heterogeneity remain central to interpreting and scaling iEEG models.

First, the practical value of learned representations depends on the preprocessing used by each system.
Strong Multi-STFT baselines remain competitive with pretrained waveform systems, particularly on datasets with smaller downstream training sets.
This provides a useful reference point for model development: engineered spectral features already support effective decoding with relatively simple models, while waveform models must learn useful features from less processed inputs.
The normalization and filtering analyses further show that apparent model advantages can depend strongly on preprocessing choices.
Evaluating learned representations against baselines under matched preprocessing is therefore essential for establishing what the model contributes beyond its input pipeline.
Comparisons across tracks additionally assess whether these learned representations yield competitive decoding systems across institutional datasets.

Second, our scaling results show that the source of additional supervision matters for target-session decoding.
Within-session scaling produces the largest and most consistent gains, while broader supervision yields smaller, task-dependent improvements.
Several residual distribution shifts may contribute to this result.
Although iMINDBench harmonizes electrode metadata, decoding targets, and preprocessing, the source recordings still differ in acquisition referencing schemes, electrode contact sizes and geometries, anatomical coverage, amplifier characteristics, and environmental noise.
These differences can change the observed signal distribution even when datasets share a decoding target, potentially making additional institutional data difficult to exploit through pooling alone.
Our experiments do not isolate the contributions of these individual factors, but they motivate representations that are robust or adaptable to heterogeneous recordings.

\section{Limitations and Future Work}

\paragraph{Evaluation scope.}
The current evaluation covers three naturalistic movie-watching datasets and assumes access to target-session calibration labels.
Extending the benchmark to other stimulus paradigms and evaluation without target-session labels would test broader forms of generalization.

\paragraph{Dataset harmonization.}
Dataset harmonization is not fully automated: incorporating a new dataset can require manual alignment of electrode locations, brain-area labels, and other metadata, alongside dataset-specific stimulus processing.
Automated alignment and validation would reduce this curation effort while preserving a consistent evaluation protocol.

\paragraph{Scaling coverage.}
The within-dataset and multi-dataset scaling experiments currently evaluate a single pretrained model, whose implementation supports joint supervised fine-tuning across subjects.
The other pretrained systems are evaluated using linear decoding heads over output embeddings, without a corresponding mechanism for joint cross-subject fine-tuning in our implementation.
Extending these evaluations to additional models would test how broadly the observed scaling pattern holds.

\paragraph{Model tuning.}
Our model comparisons also reflect the training recipes and tuning budgets used in this study; further optimization of pretraining and fine-tuning could change their relative performance.
Appendix~\ref{app:autoresearch} explores this issue through a human-guided AutoResearch-style feasibility study, in which tuning a pretraining recipe improves downstream decoding.
Such workflows could help establish stronger model baselines, provided that tuning budgets and validation-based selection procedures are documented consistently across models.

\section{Conclusion}

iMINDBench provides a shared framework for evaluating iEEG decoding across tasks, institutions, and preprocessing tracks.
Our results show that model advantages depend on the input representation and evaluation setting, while additional supervised data from other subjects and institutions do not reliably improve target-session decoding in our scaling experiments.
Progress on this benchmark will require models that extract useful features under limited supervision and learn more effectively from heterogeneous recordings, with representations robust or adaptable to differences in anatomy, acquisition, and recording conditions.
Matched-preprocessing comparisons and cross-institution evaluations provide complementary ways to measure these advances toward general-purpose iEEG decoding.

\FloatBarrier
\clearpage

{
\small
\bibliographystyle{unsrt}
\bibliography{references}

@misc{han2026diver1,
  author    = {Han, Danny Dongyeop and Gwon, Yonghyeon and Lee, Ahhyun Lucy and Lee, Taeyang and Lee, Seong Jin and Choi, Jubin and Lee, Sebin and Bang, Jihyun and Lee, Seungju and Park, David Keetae and Yoo, Shinjae and Chung, Chun Kee and Cha, Jiook},
  title     = {{DIVER-1}: Scaling Intracranial {EEG} Foundation Models for Transferable Representations},
  year      = {2026},
  eprint    = {2512.19097v3},
  archivePrefix = {arXiv},
  url       = {https://arxiv.org/abs/2512.19097v3},
  note      = {arXiv preprint}
}

@inproceedings{chau_wang_2025_population,
  title     = {Population Transformer: Learning Population-level Representations of Neural Activity},
  url       = {https://openreview.net/forum?id=FVuqJt3c4L},
  booktitle = {The Thirteenth International Conference on Learning Representations},
  author    = {Chau, Geeling and Wang, Christopher and Talukder, Sabera J and Subramaniam, Vighnesh and Soedarmadji, Saraswati and Yue, Yisong and Katz, Boris and Barbu, Andrei},
  year      = {2025}
}

@inproceedings{patel2025timescale,
  author    = {Patel, Eshani and Yue, Yisong and Chau, Geeling},
  title     = {Learning Time-Scale Invariant Population-Level Neural Representations},
  booktitle = {NeurIPS 2025 Workshop on Foundation Models for the Brain and Body},
  year      = {2025},
  eprint    = {2511.13022},
  archivePrefix = {arXiv},
  doi       = {10.48550/arXiv.2511.13022},
  url       = {https://openreview.net/forum?id=gPw0qPn908}
}

@inproceedings{wang2023brainbert,
  title     = {{BrainBERT}: Self-supervised representation learning for intracranial recordings},
  url       = {https://openreview.net/forum?id=xmcYx_reUn6},
  booktitle = {International Conference on Learning Representations},
  author    = {Wang, Christopher and Subramaniam, Vighnesh and Yaari, Adam Uri and Kreiman, Gabriel and Katz, Boris and Cases, Ignacio and Barbu, Andrei},
  year      = {2023}
}

@inproceedings{oganesian2025barista,
  author    = {Oganesian, Lucine L. and Hashemi, Saba and Shanechi, Maryam M.},
  title     = {{BaRISTA}: Brain Scale Informed Spatiotemporal Representation of Human Intracranial Neural Activity},
  booktitle = {The Thirty-ninth Conference on Neural Information Processing Systems},
  year      = {2025},
  url       = {https://openreview.net/forum?id=LDjBDk3Czb}
}

@article{peterson2021htnet,
  author    = {Peterson, Steven M. and Steine-Hanson, Zoe and Davis, Nathan and Rao, Rajesh P. N. and Brunton, Bingni W.},
  title     = {Generalized neural decoders for transfer learning across participants and recording modalities},
  journal   = {Journal of Neural Engineering},
  volume    = {18},
  number    = {2},
  pages     = {026014},
  year      = {2021},
  month     = {mar},
  publisher = {IOP Publishing},
  url       = {https://doi.org/10.1088/1741-2552/abda0b},
  doi       = {10.1088/1741-2552/abda0b}
}

@article{lawhern2018eegnet,
  title     = {{EEGNet}: a compact convolutional neural network for {EEG}-based brain--computer interfaces},
  author    = {Lawhern, Vernon J. and Solon, Amelia J. and Waytowich, Nicholas R. and Gordon, Stephen M. and Hung, Chou P. and Lance, Brent J.},
  journal   = {Journal of Neural Engineering},
  volume    = {15},
  number    = {5},
  pages     = {056013},
  year      = {2018},
  publisher = {IOP Publishing},
  doi       = {10.1088/1741-2552/aace8c}
}

@inproceedings{zhang2023brant,
  author    = {Zhang, Daoze and Yuan, Zhizhang and Yang, Yang and Chen, Junru and Wang, Jingjing and Li, Yafeng},
  title     = {{Brant}: Foundation Model for Intracranial Neural Signal},
  booktitle = {Advances in Neural Information Processing Systems},
  volume    = {36},
  pages     = {26304--26321},
  publisher = {Curran Associates, Inc.},
  year      = {2023},
  url       = {https://proceedings.neurips.cc/paper_files/paper/2023/hash/535915d26859036410b0533804cee788-Abstract-Conference.html}
}

@article{nejedly2020multicenter,
  author  = {Nejedly, Petr and Kremen, Vaclav and Sladky, Vladimir and Cimbalnik, Jan and Klimes, Petr and Plesinger, Filip and Mivalt, Filip and Travnicek, Vojtech and Viscor, Ivo and Pail, Martin and Halamek, Josef and Brinkmann, Benjamin H. and Brazdil, Milan and Jurak, Pavel and Worrell, Gregory},
  title   = {Multicenter intracranial {EEG} dataset for classification of graphoelements and artifactual signals},
  journal = {Scientific Data},
  volume  = {7},
  number  = {179},
  year    = {2020},
  doi     = {10.1038/s41597-020-0532-5},
  url     = {https://doi.org/10.1038/s41597-020-0532-5}
}

@inproceedings{duan2026omniieeg,
  author    = {Duan, Chenda and Zhang, Yipeng and Kanai, Sotaro and Ding, Yuanyi and Daida, Atsuro and Yu, Pengyue and Zheng, Tiancheng and Kuroda, Naoto and Hussain, Shaun A. and Asano, Eishi and Nariai, Hiroki and Roychowdhury, Vwani},
  title     = {{Omni-iEEG}: A Large-Scale, Comprehensive {iEEG} Dataset and Benchmark for Epilepsy Research},
  booktitle = {International Conference on Learning Representations},
  year      = {2026},
  url       = {https://openreview.net/forum?id=rv9lQpY5cG}
}

@misc{zahorodnii2025neuroprobe,
  author        = {Zahorodnii, Andrii and Wang, Christopher and Stankovits, Bennett and Moraitaki, Charikleia and Chau, Geeling and Barbu, Andrei and Katz, Boris and Fiete, Ila R.},
  title         = {Neuroprobe: Evaluating Intracranial Brain Responses to Naturalistic Stimuli},
  year          = {2025},
  eprint        = {2509.21671},
  archivePrefix = {arXiv},
  primaryClass  = {cs.LG},
  url           = {https://arxiv.org/abs/2509.21671},
  note          = {arXiv preprint}
}

@inproceedings{wang2024braintreebank,
  author    = {Wang, Christopher and Yaari, Adam Uri and Singh, Aaditya K. and Subramaniam, Vighnesh and Rosenfarb, Dana and DeWitt, Jan and Misra, Pranav and Madsen, Joseph R. and Stone, Scellig and Kreiman, Gabriel and Katz, Boris and Cases, Ignacio and Barbu, Andrei},
  title     = {Brain Treebank: Large-scale intracranial recordings from naturalistic language stimuli},
  booktitle = {Advances in Neural Information Processing Systems},
  volume    = {37},
  year      = {2024},
  doi       = {10.52202/079017-3060},
  url       = {https://proceedings.neurips.cc/paper_files/paper/2024/hash/aefa2385b3f33abf1526ae4e2c208cd9-Abstract-Datasets_and_Benchmarks_Track.html}
}

@article{keles2024multimodal,
  author  = {Keles, Umit and Dubois, Julien and Le, Kevin J. M. and Tyszka, J. Michael and Kahn, David A. and Reed, Chrystal M. and Chung, Jeffrey M. and Mamelak, Adam N. and Adolphs, Ralph and Rutishauser, Ueli},
  title   = {Multimodal single-neuron, intracranial {EEG}, and {fMRI} brain responses during movie watching in human patients},
  journal = {Scientific Data},
  volume  = {11},
  number  = {214},
  year    = {2024},
  doi     = {10.1038/s41597-024-03029-1},
  url     = {https://doi.org/10.1038/s41597-024-03029-1}
}

@article{berezutskaya2022open,
  author  = {Berezutskaya, Julia and Vansteensel, Mariska J. and Aarnoutse, Erik J. and Freudenburg, Zachary V. and Piantoni, Giovanni and Branco, Mariana P. and Ramsey, Nick F.},
  title   = {Open multimodal {iEEG-fMRI} dataset from naturalistic stimulation with a short audiovisual film},
  journal = {Scientific Data},
  volume  = {9},
  number  = {91},
  year    = {2022},
  doi     = {10.1038/s41597-022-01173-0},
  url     = {https://doi.org/10.1038/s41597-022-01173-0}
}

@article{peterson2022ajile12,
  author  = {Peterson, Steven M. and Singh, Satpreet H. and Dichter, Benjamin and Scheid, Michael and Rao, Rajesh P. N. and Brunton, Bingni W.},
  title   = {{AJILE12}: Long-term naturalistic human intracranial neural recordings and pose},
  journal = {Scientific Data},
  volume  = {9},
  number  = {184},
  year    = {2022},
  doi     = {10.1038/s41597-022-01280-y},
  url     = {https://doi.org/10.1038/s41597-022-01280-y}
}

@article{jayaram2018moabb,
  author    = {Jayaram, Vinay and Barachant, Alexandre},
  title     = {{MOABB}: trustworthy algorithm benchmarking for {BCIs}},
  journal   = {Journal of Neural Engineering},
  volume    = {15},
  number    = {6},
  pages     = {066011},
  year      = {2018},
  doi       = {10.1088/1741-2552/aadea0},
  url       = {https://doi.org/10.1088/1741-2552/aadea0}
}

@article{lachaux2012high,
  author  = {Lachaux, Jean-Philippe and Axmacher, Nikolai and Mormann, Florian and Halgren, Eric and Crone, Nathan E.},
  title   = {High-frequency neural activity and human cognition: Past, present and possible future of intracranial {EEG} research},
  journal = {Progress in Neurobiology},
  volume  = {98},
  number  = {3},
  pages   = {279--301},
  year    = {2012},
  doi     = {10.1016/j.pneurobio.2012.06.008},
  url     = {https://doi.org/10.1016/j.pneurobio.2012.06.008}
}

@article{parvizi2018promises,
  author  = {Parvizi, Josef and Kastner, Sabine},
  title   = {Promises and limitations of human intracranial electroencephalography},
  journal = {Nature Neuroscience},
  volume  = {21},
  pages   = {474--483},
  year    = {2018},
  doi     = {10.1038/s41593-018-0108-2},
  url     = {https://doi.org/10.1038/s41593-018-0108-2}
}

@inproceedings{azabou2023unified,
  author    = {Azabou, Mehdi and Arora, Vinam and Ganesh, Venkataramana and Mao, Ximeng and Nachimuthu, Santosh and Mendelson, Michael and Richards, Blake and Perich, Matthew and Lajoie, Guillaume and Dyer, Eva L.},
  title     = {A Unified, Scalable Framework for Neural Population Decoding},
  booktitle = {Thirty-seventh Conference on Neural Information Processing Systems},
  year      = {2023},
  url       = {https://papers.nips.cc/paper_files/paper/2023/hash/8ca113d122584f12a6727341aaf58887-Abstract-Conference.html}
}

@inproceedings{jiang2024labram,
  author    = {Jiang, Weibang and Zhao, Liming and Lu, Bao-liang},
  title     = {Large Brain Model for Learning Generic Representations with Tremendous {EEG} Data in {BCI}},
  booktitle = {The Twelfth International Conference on Learning Representations},
  year      = {2024},
  url       = {https://openreview.net/forum?id=QzTpTRVtrP}
}

@inproceedings{carzaniga2026mvpformer,
  author    = {Carzaniga, Francesco S. and Hersche, Michael and Sebastian, Abu and Schindler, Kaspar and Rahimi, Abbas},
  title     = {A Foundation Model with Multi-variate Parallel Attention to Generate Neuronal Activity},
  booktitle = {The Fourteenth International Conference on Learning Representations},
  year      = {2026},
  url       = {https://openreview.net/forum?id=5M1YOW3bRq}
}

@inproceedings{wang2024eegpt,
  author    = {Wang, Guangyu and Liu, Wenchao and He, Yuhong and Xu, Cong and Ma, Lin and Li, Haifeng},
  title     = {{EEGPT}: Pretrained Transformer for Universal and Reliable Representation of {EEG} Signals},
  booktitle = {Advances in Neural Information Processing Systems},
  volume    = {37},
  year      = {2024},
  url       = {https://proceedings.neurips.cc/paper_files/paper/2024/hash/4540d267eeec4e5dbd9dae9448f0b739-Abstract-Conference.html}
}

@inproceedings{kuruppu2025eegreview,
  author    = {Kuruppu, Gayal and Wagh, Neeraj and Varatharajah, Yogatheesan},
  title     = {{EEG} Foundation Models: A Critical Review of Current Progress and Future Directions},
  booktitle = {NeurIPS 2025 Workshop on BrainBodyFM},
  year      = {2025},
  url       = {https://openreview.net/forum?id=Iu6qVgtgUD}
}

@inproceedings{yang2026eegworth,
  author    = {Yang, Liuyin and Sun, Qiang and Li, Ang and Van Hulle, Marc M.},
  title     = {Are {EEG} Foundation Models Worth It? Comparative Evaluation with Traditional Decoders in Diverse {BCI} Tasks},
  booktitle = {The Fourteenth International Conference on Learning Representations},
  year      = {2026},
  url       = {https://openreview.net/forum?id=5Xwm8e6vbh}
}

@misc{karpathy2026autoresearch,
  author       = {Karpathy, Andrej},
  title        = {autoresearch: {AI} Agents Running Research on Single-{GPU} Nanochat Training Automatically},
  year         = {2026},
  howpublished = {\url{https://github.com/karpathy/autoresearch}},
  note         = {GitHub repository}
}

@article{li2018optimal,
  title={Optimal referencing for stereo-electroencephalographic (SEEG) recordings},
  author={Li, Guangye and Jiang, Shize and Paraskevopoulou, Sivylla E and Wang, Meng and Xu, Yang and Wu, Zehan and Chen, Liang and Zhang, Dingguo and Schalk, Gerwin},
  journal={NeuroImage},
  volume={183},
  pages={327--335},
  year={2018},
  publisher={Elsevier}
}

@article{buzsaki2012origin,
  title={The origin of extracellular fields and currents—EEG, ECoG, LFP and spikes},
  author={Buzs{\'a}ki, Gy{\"o}rgy and Anastassiou, Costas A and Koch, Christof},
  journal={Nature reviews neuroscience},
  volume={13},
  number={6},
  pages={407--420},
  year={2012},
  publisher={Nature Publishing Group UK London}
}

@article{baevski2020wav2vec,
  title={wav2vec 2.0: A framework for self-supervised learning of speech representations},
  author={Baevski, Alexei and Zhou, Yuhao and Mohamed, Abdelrahman and Auli, Michael},
  journal={Advances in neural information processing systems},
  volume={33},
  pages={12449--12460},
  year={2020}
}
}

\clearpage
\appendix
\raggedbottom
\section*{Appendix}

\section{Dataset Harmonization and Acquisition Metadata}
\label{app:dataset-harmonization}

Controlled cross-institution evaluation requires alignment along two benchmark-specific dimensions: decoding labels derived from independently collected movie stimuli and iEEG electrode metadata reported under different localization conventions.
Together, these steps support common decoding targets and spatial metadata without treating the source datasets as uniform.

\paragraph{Task and temporal alignment.}
For the BYD and Pippi components, task and label alignment starts from the Neuroprobe/Brain Treebank stimulus-feature extraction pipeline, applied to each dataset's video stimulus where source annotations support it.
Task labels, scalar thresholds, and lexical/nonverbal row construction are defined in Appendix~\ref{app:decoding-tasks}.
Sentence-level linguistic attributes are labeled using the Brain Treebank annotation protocol, and sentence and word onset times are initialized with Whisper automatic speech recognition (ASR) before manual alignment using 4$\times$ slowed video playback.
We manually correct \texttt{face\_num} labels because the automatic face-count detector is noisy on these stimuli.

\paragraph{Electrode metadata alignment.}
Electrode layouts and cortical coverage vary by patient, and localization metadata use different coordinate systems and anatomical labels.
Electrode locations are transformed into Brain Treebank's left-posterior-inferior (LPI) coordinate space by first aligning coordinate dimensions to a sensible brain orientation and then matching the electrode coordinate ranges provided in Brain Treebank.
For brain-area metadata, we harmonize labels to the Destrieux parcellation used by Brain Treebank: Pippi electrodes are overlaid onto Destrieux areas with a FreeSurfer alignment pipeline using the publicly available MRI scans, and BYD electrode labels are manually converted to the same Destrieux area set.

\paragraph{Remaining acquisition differences.}
Table~\ref{tab:dataset-metadata} summarizes the remaining acquisition and stimulus differences relevant to multi-dataset model transfer, using metadata reported in the source dataset papers.

\begin{table}[h]
  \centering
  \footnotesize
  \setlength{\tabcolsep}{3pt}
  \caption{\textbf{Institution-specific stimulus and acquisition metadata.}
  Dataset components differ in stimulus duration and language, electrode type and coverage, sampling, reference conventions, and coordinate reporting.
  BYD and Pippi entries focus on macroelectrode and clinical sEEG recordings, respectively.}
  \label{tab:dataset-metadata}
  \renewcommand{\arraystretch}{1.12}

  \begin{tabularx}{\linewidth}{@{}>{\raggedright\arraybackslash}p{0.14\linewidth}*{3}{>{\raggedright\arraybackslash}X}@{}}
    \toprule
    \textbf{Property}
      & \textbf{Neuroprobe / Brain Treebank}~\cite{zahorodnii2025neuroprobe,wang2024braintreebank}
      & \textbf{BYD / Cedars-Sinai}~\cite{keles2024multimodal}
      & \textbf{Pippi / Utrecht}~\cite{berezutskaya2022open} \\
    \midrule
    \textbf{Stimulus}
      & English feature-length Hollywood films. Brain Treebank reports 26 films and 43.5 total hours; subjects watched 2.6 movies on average.
      & 8-minute black-and-white Hitchcock excerpt with English audio-visual content.
      & 6.5-minute Dutch-dubbed short film from \emph{Pippi on the Run}, with interleaved speech and music blocks. \\
    \addlinespace
    \textbf{Electrodes}
      & sEEG depth probes with 6--16 contacts per probe. Contacts: 0.8 mm diameter, 2 mm long.
      & Macroelectrodes on hybrid Behnke-Fried depth electrodes.
      & Clinical sEEG recordings. \\
    \addlinespace
    \textbf{Sampling}
      & 2048 Hz.
      & 1000 Hz.
      & 2048 Hz. \\
    \addlinespace
    \textbf{Source filtering}
      & 60 Hz notch filtering and harmonics.
      & 60 Hz notch and 0.1 Hz high-pass filtering.
      & Source analysis applies 50 Hz notch filtering. \\
    \addlinespace
    \textbf{Referencing}
      & Neuroprobe evaluates Laplacian-rereferenced inputs for several benchmark models.
      & Source processing uses common-average rereferencing.
      & External mastoid reference during acquisition; common-average rereferencing in source analysis. \\
    \addlinespace
    \textbf{Localization}
      & Electrode positions localized to an average cortical atlas.
      & Native-space and MNI152 coordinates provided.
      & Native-space electrode locations; MNI projection used for visualization. \\
    \bottomrule
  \end{tabularx}
  \par\smallskip
  \raggedright
  sEEG: stereoelectroencephalography;
  MNI: Montreal Neurological Institute.
\end{table}

\FloatBarrier

\section{Decoding Tasks}
\label{app:decoding-tasks}

Task definitions follow the Neuroprobe task suite~\cite{zahorodnii2025neuroprobe} and Population Transformer benchmark protocol~\cite{chau_wang_2025_population} where source annotations support them.
For scalar binary tasks, Neuroprobe uses bottom-versus-top quartiles, while BYD and Pippi use bottom-versus-top terciles to improve class support in the shorter stimuli.
Lexical and surprisal tasks are restricted to word rows.

For BYD and Pippi visual and auditory tasks, feature tables include lexical word rows and sampled nonverbal rows.
Nonverbal rows are sampled in 500 ms windows stepped every 500 ms from gaps longer than 2.0 s between the preceding word's offset and the next word's onset.
They inherit the preceding word row's sentence identifier and provide the negative class for speech and sentence-onset tasks.

\begin{table}[H]
  \centering
  \scriptsize
  \setlength{\tabcolsep}{3pt}
  \renewcommand{\arraystretch}{1.02}
  \caption{\textbf{Aligned visual, auditory, and language decoding tasks.}
  All tasks use binary classification on 1-second neural windows aligned to word onset or sampled nonverbal-interval onset.
  Classes are rebalanced before training.}
  \label{tab:decoding-tasks}
  \begin{tabular}{@{}>{\raggedright\arraybackslash}p{0.04\linewidth}>{\raggedright\arraybackslash}p{0.20\linewidth}>{\raggedright\arraybackslash}p{0.43\linewidth}>{\raggedright\arraybackslash}p{0.25\linewidth}@{}}
    \toprule
    \textbf{\#} & \textbf{Feature} & \textbf{Description} & \textbf{Benchmark task} \\
    \midrule
    1 & \texttt{onset} \emph{(language)}
      & Whether a new sentence starts.
      & Sentence onset vs. nonverbal interval. \\
    2 & \texttt{speech} \emph{(language)}
      & Whether speech is present.
      & Word interval vs. nonverbal interval. \\
    3 & \texttt{word\_index} \emph{(language)}
      & Word index within its sentence.
      & First word ($0$) vs. second word ($1$); later words excluded. \\
    4 & \texttt{word\_gap} \emph{(language)}
      & Time from the preceding word's offset to the current word's onset within the same sentence (ms).
      & Low vs. high value. \\
    5 & \texttt{gpt2\_surprisal} \emph{(language)}
      & Negative log probability of the word under GPT-2, conditioned on the preceding 20 s of language context.
      & Low vs. high value. \\
    6 & \texttt{word\_head\_pos} \emph{(language)}
      & Relative position of the word's dependency-tree head.
      & Head to the left/root/self vs. head to the right. \\
    7 & \texttt{word\_length} \emph{(language)}
      & Word duration in milliseconds.
      & Low vs. high value. \\
    8 & \texttt{word\_part\_speech} \emph{(language)}
      & Universal part-of-speech (UPOS) tag of the word.
      & Noun vs. verb; other tags excluded. \\
    9 & \texttt{delta\_volume} \emph{(auditory)}
      & Difference in average RMS audio power between the 500 ms windows before and after target onset.
      & Low vs. high value. \\
    10 & \texttt{volume} \emph{(auditory)}
      & Average root mean square (RMS) audio power.
      & Low vs. high value. \\
    11 & \texttt{pitch} \emph{(auditory)}
      & Average audio pitch.
      & Low vs. high value. \\
    12 & \texttt{global\_flow} \emph{(visual)}
      & Camera-motion proxy computed as the maximal average dense optical-flow vector magnitude.
      & Low vs. high value. \\
    13 & \texttt{local\_flow} \emph{(visual)}
      & Large-displacement proxy computed as the maximal individual optical-flow vector magnitude.
      & Low vs. high value. \\
    14 & \texttt{frame\_brightness} \emph{(visual)}
      & Mean HSV value (V) over pixels.
      & Low vs. high value. \\
    15 & \texttt{face\_num} \emph{(visual)}
      & Maximum number of faces per frame during the target interval.
      & No face ($0$) vs. at least one face ($\geq 1$). \\
    \bottomrule
  \end{tabular}
\end{table}

\FloatBarrier

\clearpage
\section{Additional Unit Decodability Subset Scorecards}
\label{app:subject-subset-scorecards}

Table~\ref{tab:scoreboard-all-units} reports the same model and track rows as the Main scorecard on the All subset.
Table~\ref{tab:scoreboard-challenge-units} reports the same rows on the Challenge subset.

\begin{table}[H]
  \centering
  \normalsize
  \setlength{\tabcolsep}{4pt}
  \renewcommand{\arraystretch}{0.9}
  \begin{tabular}{@{}lllcccc@{}}
    \toprule
    \textbf{Track} & Model & Pretraining & Overall & Neuroprobe & BYD & Pippi \\
    \midrule
    \multicolumn{7}{l}{\textbf{Multi-STFT}} \\
     & Logistic & not pretrained & \aucCell{26}{0.587} & \aucCell{15}{0.641} & \aucCell{31}{0.576} & \aucCell{44}{0.545} \\
     & MLP & not pretrained & \aucCell{37}{\underline{0.592}} & \aucCell{26}{0.646} & \aucCell{36}{\underline{0.578}} & \aucCell{53}{\underline{0.551}} \\
     & CNN & not pretrained & \aucCell{15}{0.583} & \aucCell{44}{\underline{0.654}} & \aucCell{15}{0.570} & \aucCell{15}{0.525} \\
     & PopT-v2 & BrainTreeBank & \aucCell{55}{\textbf{0.599}} & \aucCell{55}{\textbf{0.659}} & \aucCell{55}{\textbf{0.585}} & \aucCell{55}{\textbf{0.553}} \\
    \midrule
    \multicolumn{7}{l}{\textbf{Waveform}} \\
     & Logistic & not pretrained & \aucCell{15}{0.541} & \aucCell{15}{0.589} & \aucCell{15}{0.515} & \aucCell{15}{0.520} \\
     & MLP & not pretrained & \aucCell{25}{0.554} & \aucCell{27}{0.612} & \aucCell{19}{0.523} & \aucCell{27}{0.527} \\
     & CNN & not pretrained & \aucCell{25}{0.554} & \aucCell{29}{0.616} & \aucCell{20}{0.524} & \aucCell{19}{0.523} \\
     & HTNet & not pretrained & \aucCell{40}{0.574} & \aucCell{40}{0.636} & \aucCell{39}{0.556} & \aucCell{31}{0.530} \\
     & BaRISTA & BrainTreeBank & \aucCell{54}{\underline{0.593}} & \aucCell{50}{\underline{0.655}} & \aucCell{55}{\textbf{0.583}} & \aucCell{46}{\underline{0.540}} \\
     & DIVER-1 & private+ AJILE12 & \aucCell{55}{\textbf{0.593}} & \aucCell{55}{\textbf{0.666}} & \aucCell{47}{\underline{0.569}} & \aucCell{55}{\textbf{0.545}} \\
    \bottomrule
\end{tabular}
\caption{\textbf{All benchmark results.}
Within-session ROC-AUC performance on all supported units, grouped by the same fixed-track and model-native families as Table~\ref{tab:scoreboard}.
Cells are shaded with a normalized color scheme within column and track.
Best per column and track are bolded, and second-best values are underlined.}
  \label{tab:scoreboard-all-units}
\end{table}

\begin{table}[H]
  \centering
  \normalsize
  \setlength{\tabcolsep}{4pt}
  \renewcommand{\arraystretch}{0.9}
  \begin{tabular}{@{}lllcccc@{}}
    \toprule
    \textbf{Track} & Model & Pretraining & Overall & Neuroprobe & BYD & Pippi \\
    \midrule
    \multicolumn{7}{l}{\textbf{Multi-STFT}} \\
     & Logistic & not pretrained & \aucCell{15}{0.521} & \aucCell{15}{0.536} & \aucCell{15}{0.509} & \aucCell{37}{\underline{0.519}} \\
     & MLP & not pretrained & \aucCell{55}{\textbf{0.527}} & \aucCell{30}{\underline{0.541}} & \aucCell{23}{0.512} & \aucCell{55}{\textbf{0.529}} \\
     & CNN & not pretrained & \aucCell{24}{0.523} & \aucCell{55}{\textbf{0.550}} & \aucCell{26}{\underline{0.512}} & \aucCell{15}{0.506} \\
     & PopT-v2 & BrainTreeBank & \aucCell{41}{\underline{0.525}} & \aucCell{25}{0.539} & \aucCell{55}{\textbf{0.521}} & \aucCell{31}{0.515} \\
    \midrule
    \multicolumn{7}{l}{\textbf{Waveform}} \\
     & Logistic & not pretrained & \aucCell{39}{0.519} & \aucCell{35}{0.535} & \aucCell{36}{0.510} & \aucCell{45}{0.513} \\
     & MLP & not pretrained & \aucCell{42}{0.520} & \aucCell{34}{0.534} & \aucCell{37}{0.510} & \aucCell{54}{\underline{0.518}} \\
     & CNN & not pretrained & \aucCell{29}{0.515} & \aucCell{15}{0.522} & \aucCell{27}{0.505} & \aucCell{55}{\textbf{0.518}} \\
     & HTNet & not pretrained & \aucCell{15}{0.509} & \aucCell{24}{0.528} & \aucCell{15}{0.500} & \aucCell{15}{0.500} \\
     & BaRISTA & BrainTreeBank & \aucCell{47}{\underline{0.523}} & \aucCell{44}{\underline{0.541}} & \aucCell{45}{\underline{0.513}} & \aucCell{45}{0.514} \\
     & DIVER-1 & private+ AJILE12 & \aucCell{55}{\textbf{0.526}} & \aucCell{55}{\textbf{0.548}} & \aucCell{55}{\textbf{0.518}} & \aucCell{41}{0.512} \\
    \bottomrule
\end{tabular}
\caption{\textbf{Challenge benchmark results.}
Within-session ROC-AUC performance on Challenge units, grouped by the same fixed-track and model-native families as Table~\ref{tab:scoreboard}.
Cells are shaded with a normalized color scheme within column and track.
Best per column and track are bolded, and second-best values are underlined.}
  \label{tab:scoreboard-challenge-units}
\end{table}

\clearpage

\section{BrainBERT Model-Native Scorecard}
\label{app:brainbert-scorecards}

Table~\ref{tab:scoreboard-brainbert-main} evaluates BrainBERT~\cite{wang2023brainbert} through its model-native single-STFT route and includes corresponding non-pretrained single-STFT decoders on the Main unit decodability subset.
We evaluate the released BrainBERT checkpoint as a frozen feature extractor with a trained linear head. Direct single-STFT baselines achieve higher overall ROC-AUC under the benchmark protocol. The evaluation uses one-second windows and training-split normalization, differing from the checkpoint's five-second pretraining windows and original normalization; it therefore characterizes the released representation under these conditions.

\begin{table}[H]
  {
  \centering
  \normalsize
  \setlength{\tabcolsep}{4pt}
  \renewcommand{\arraystretch}{0.9}
  \begin{tabular}{@{}lllcccc@{}}
    \toprule
    \textbf{Track} & Model & Pretraining & Overall & Neuroprobe & BYD & Pippi \\
    \midrule
    \multicolumn{7}{l}{\textbf{BrainBERT STFT}} \\
     & Logistic & not pretrained & \aucCell{44}{\underline{0.630}} & \aucCell{48}{0.695} & \aucCell{55}{\textbf{0.627}} & \aucCell{36}{0.569} \\
     & MLP & not pretrained & \aucCell{55}{\textbf{0.634}} & \aucCell{49}{\underline{0.696}} & \aucCell{47}{\underline{0.622}} & \aucCell{55}{\textbf{0.583}} \\
     & CNN & not pretrained & \aucCell{18}{0.620} & \aucCell{55}{\textbf{0.702}} & \aucCell{15}{0.603} & \aucCell{15}{0.553} \\
     & \shortstack[l]{BrainBERT\\(frozen + linear head)} & BrainTreeBank & \aucCell{15}{0.619} & \aucCell{15}{0.664} & \aucCell{42}{0.619} & \aucCell{44}{\underline{0.575}} \\
    \bottomrule
  \end{tabular}
  \caption{\textbf{BrainBERT model-native preprocessing scorecard on Main units.}
  Within-session ROC-AUC performance on Main units for BrainBERT and corresponding non-pretrained single-STFT decoders.
  BrainBERT is evaluated using frozen representations pretrained on BrainTreeBank, while Logistic, MLP, and CNN are fit directly on the dataset-specific single-STFT route.
  Overall is the unweighted mean of the three dataset-level scores; each dataset score is a unit-weighted mean.
  Cells are shaded within column, with the best value bolded and the second-best value underlined.}
  \label{tab:scoreboard-brainbert-main}
  }
\end{table}

\clearpage

\section{Main and Challenge Unit Coverage}
\label{app:main-challenge-unit-coverage}

\begin{figure}[H]
  \centering
  \includegraphics[width=\linewidth]{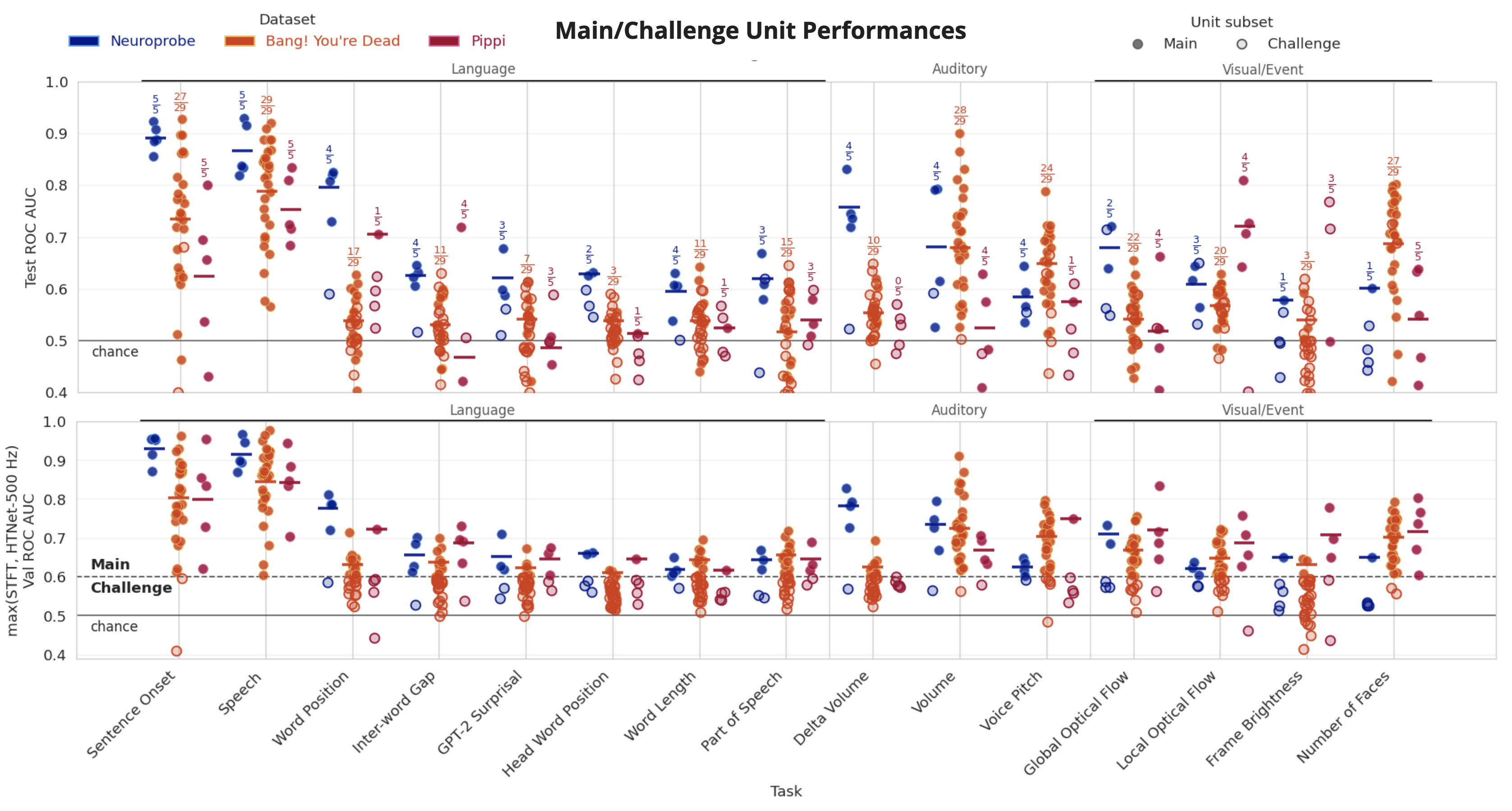}
  \caption{\textbf{Main and Challenge unit performance across tasks.}
Each point shows an evaluation unit's fold-averaged ROC-AUC.
The top panel reports test performance for the STFT screening decoder, while the bottom panel reports the validation screening score: the maximum mean validation ROC-AUC across the STFT screening decoder and HTNet at 500 Hz.
The dashed line marks the Main-unit threshold (validation ROC-AUC $>0.60$), and the solid line marks chance performance.}
  \label{fig:appendix-main-challenge-unit-performance}
\end{figure}

\begin{figure}[H]
  \centering
  \includegraphics[width=\linewidth]{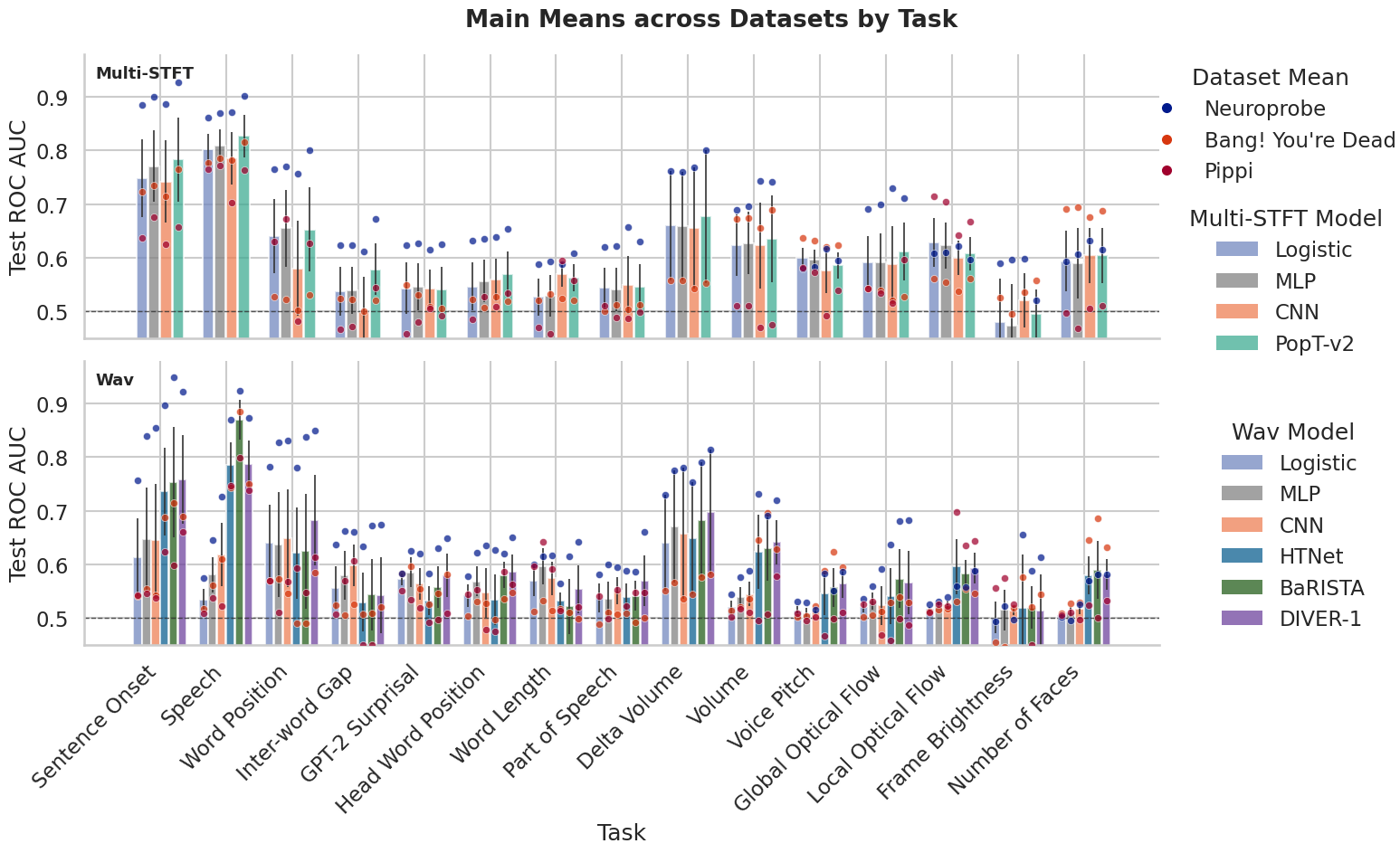}
  \caption{\textbf{Main-unit task breakout by model family.}
Bars show mean test ROC-AUC on Main units for each task and model family, with points showing dataset means.
The dashed horizontal line marks chance performance. Error bars indicate SEM across the 3 dataset means. }
  \label{fig:appendix-task-breakout-main}
\end{figure}

\begin{figure}[H]
  \centering
  \includegraphics[width=0.95\linewidth,height=0.72\textheight,keepaspectratio]{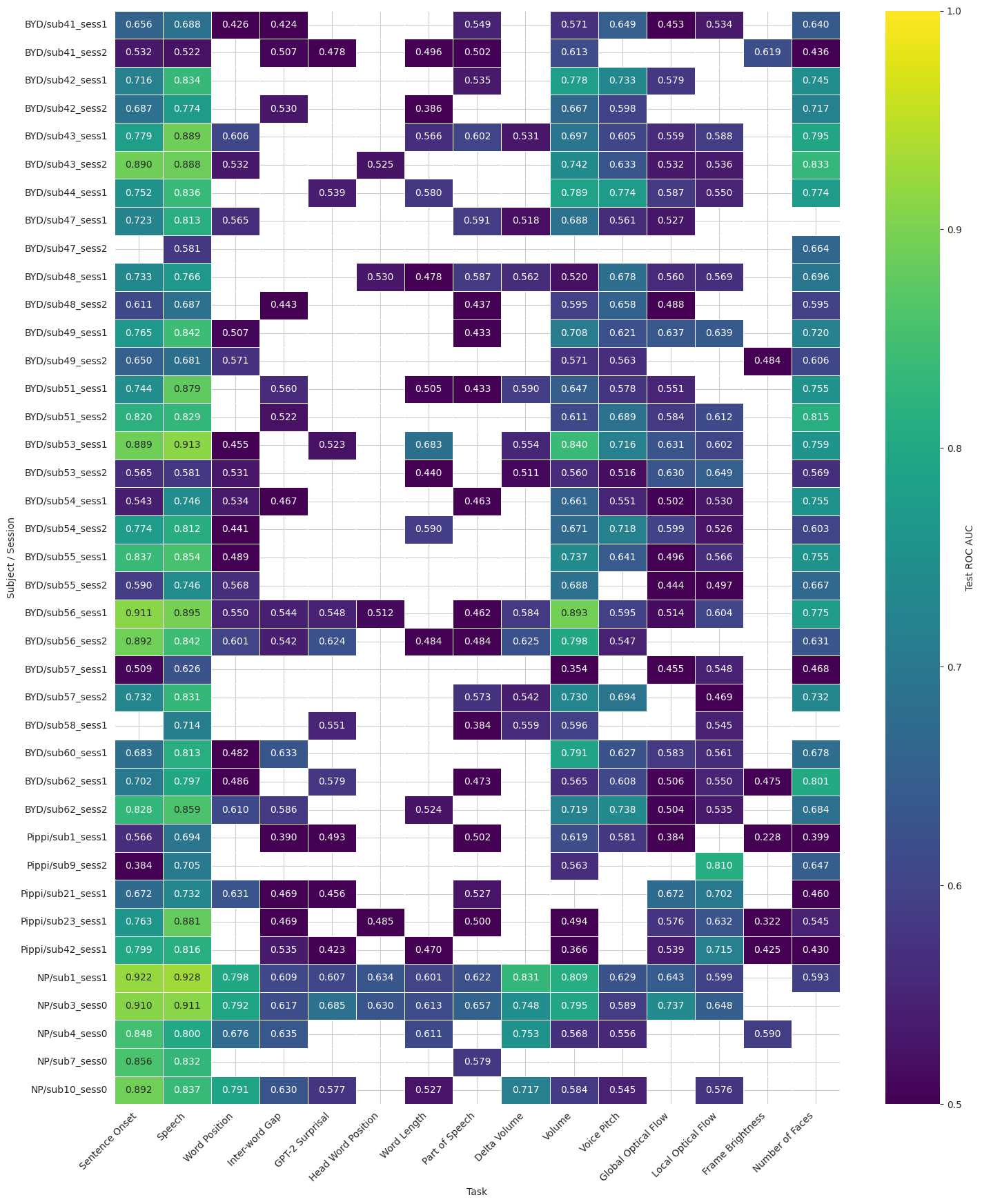}
  \caption{\textbf{Main subject/session coverage across benchmark tasks.}
Cells show fold-averaged test ROC-AUC from the STFT screening decoder for Main subject/session--task units.
Blank cells correspond to Challenge or unsupported units, and higher values indicate stronger screening-decoder test performance.}
  \label{fig:appendix-main-sub-sess}
\end{figure}

\FloatBarrier

\section{Task-Level Within-Session Scaling}
\label{app:task-sample-efficiency}

\begin{figure}[H]
  \centering
  \includegraphics[width=\linewidth,height=0.82\textheight,keepaspectratio]{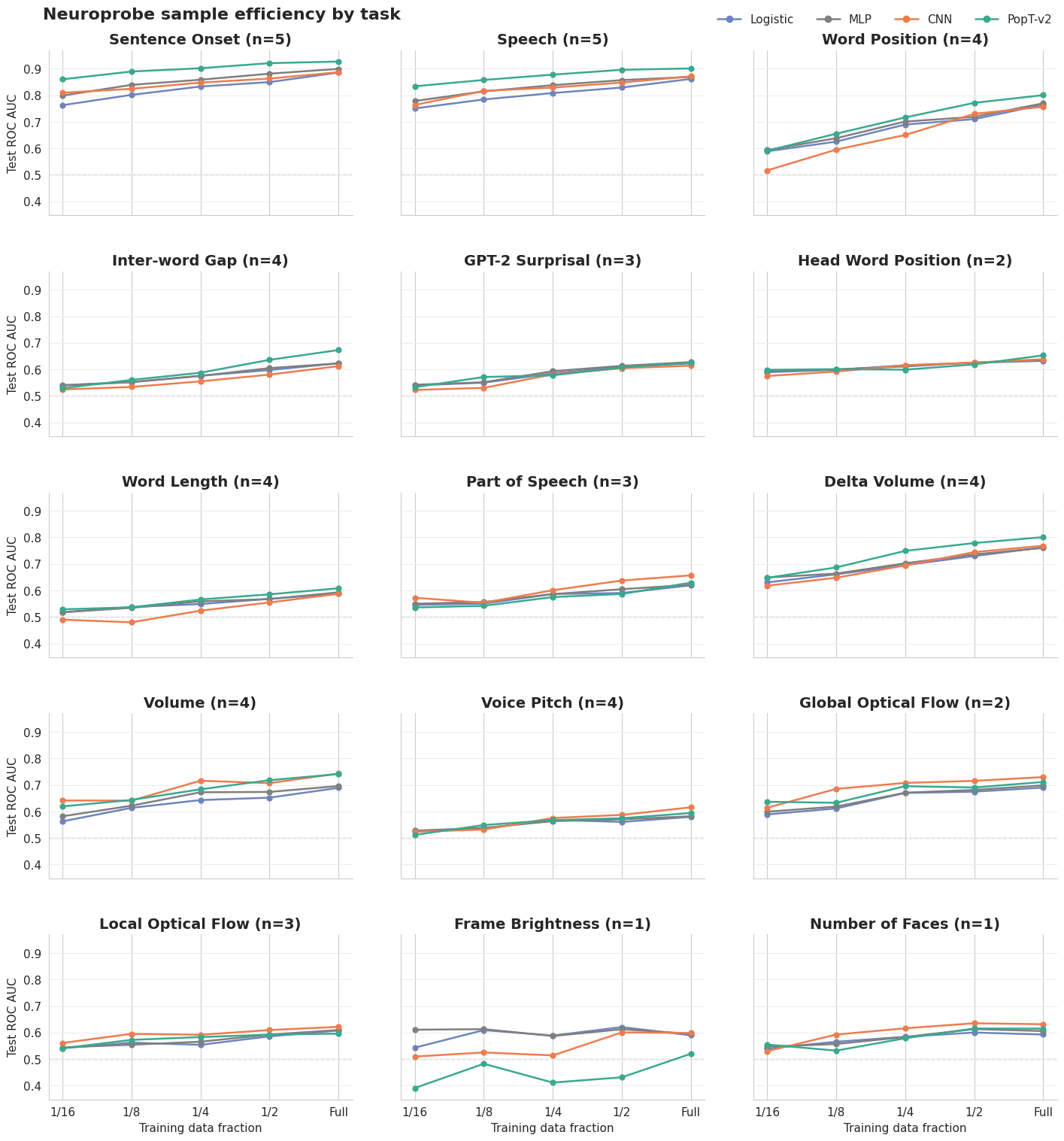}
  \caption{\textbf{Neuroprobe within-session sample efficiency by task.}
Each panel decomposes Figure~\ref{fig:data-scaling}a by decoding task and model family, showing unit-weighted mean test ROC-AUC as the available target-session training data increase from one-sixteenth to the full training set.
Only complete-case task/subject-session/fold units containing every model family and data fraction are retained.
Panel titles report the number of unique supported subjects ($n$), and the dashed horizontal line marks chance performance.}
  \label{fig:appendix-task-sample-efficiency}
\end{figure}

\FloatBarrier

\section{Task-Level Scaling Trajectories}
\label{app:task-level-scaling}

\begin{figure}[H]
  \centering
  \includegraphics[width=\linewidth]{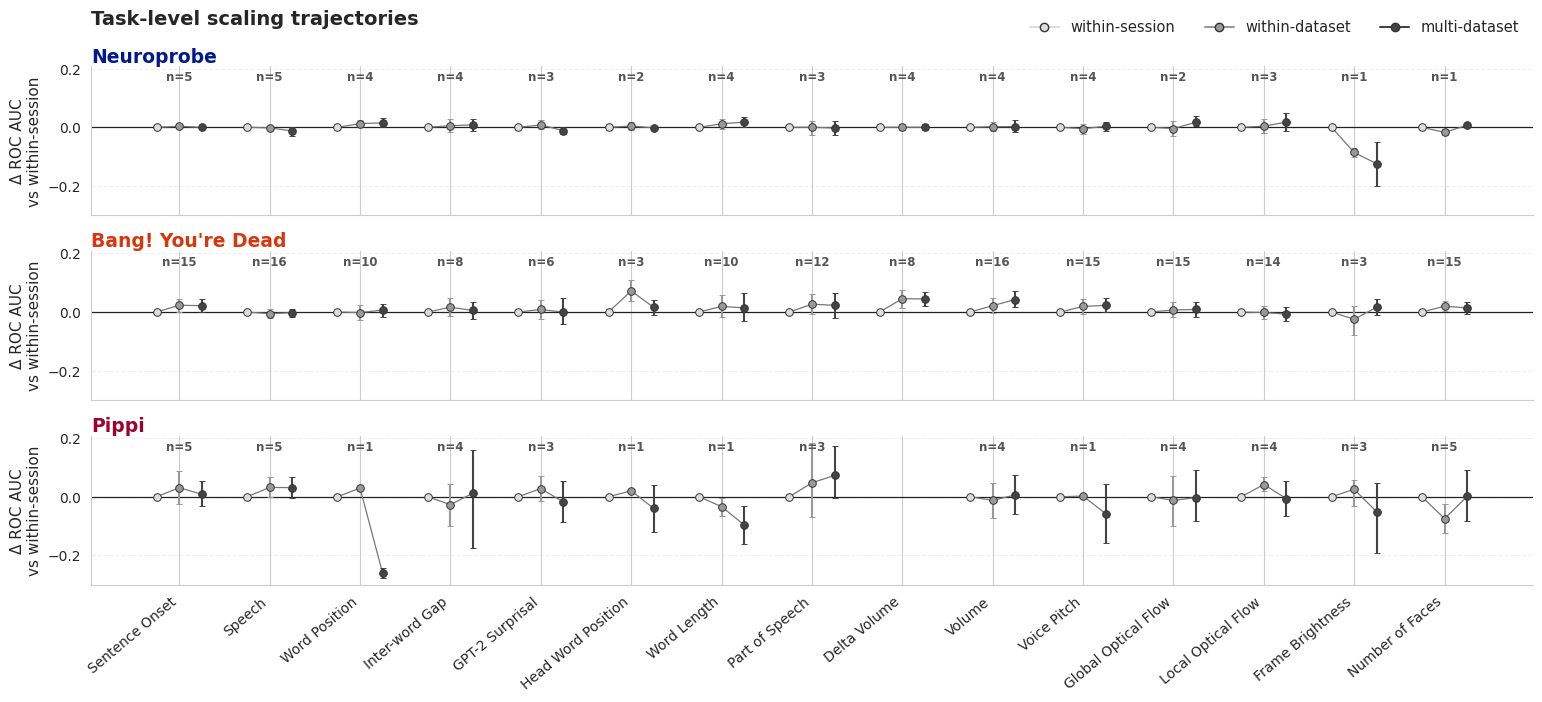}
  \caption{\textbf{Task-level scaling trajectories across datasets.}
For PopT-v2, points show the change in ROC-AUC for each decoding task relative to within-session evaluation when scaling to within-dataset and multi-dataset training; within-session values are therefore fixed at zero.
Rows correspond to Neuroprobe, Bang! You're Dead (BYD), and Pippi.
Error bars denote 95\% confidence intervals, and labels above each task report the number of unique supported subjects ($n$).
Positive values indicate improvement over within-session training, while negative values indicate reduced performance.}
  \label{fig:appendix-task-scaling}
\end{figure}

\clearpage

\section{Multi-STFT Input Example}
\label{app:multistft-example}

\begin{figure}[H]
  \centering
  \includegraphics[width=\linewidth]{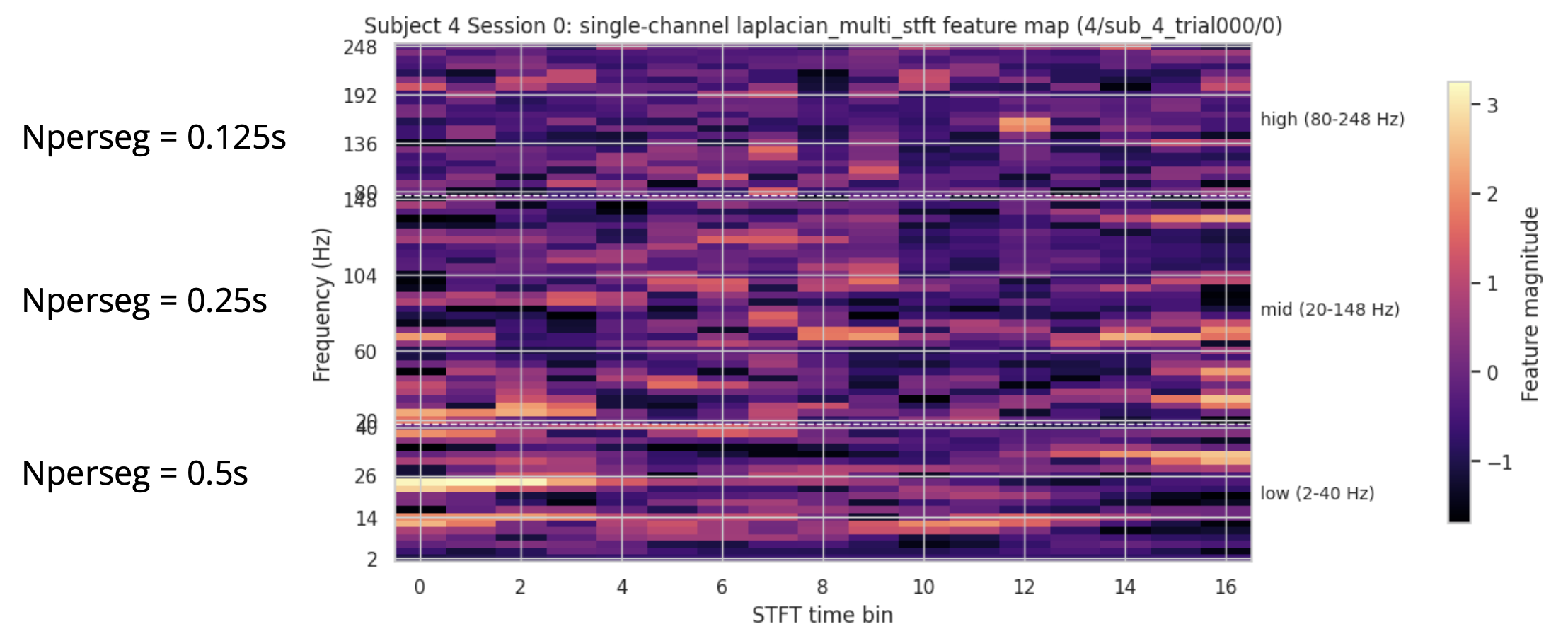}
  \caption{\textbf{Example Multi-STFT input representation.}
The low-, mid-, and high-frequency sections use 0.5-, 0.25-, and 0.125-second windows with an approximately 62-ms hop and target 2--40, 20--150, and 80--250 Hz, respectively.
The sections are concatenated along frequency with aligned time bins.
Colors show feature magnitudes standardized per channel and frequency bin using training-split statistics.}
  \label{fig:appendix-multistft}
\end{figure}

\begin{table}[H]
  \centering
  \small
  \begin{tabular}{@{}lrrr@{}}
    \toprule
    Setting & Low & Mid & High \\
    \midrule
    Window samples at 2048 Hz & 1024 & 512 & 256 \\
    Window samples at 1000 Hz (BYD) & 500 & 250 & 125 \\
    Retained upper frequency at 2048 Hz (Hz) & 40 & 148 & 248 \\
    \bottomrule
  \end{tabular}
  \caption{\textbf{Exact Multi-STFT settings.}
  Window sample counts specify \texttt{nperseg}; hops are 128 samples at 2048 Hz and 62 samples at 1000 Hz.
  Retained upper frequencies reflect discrete-bin spacing.}
  \label{tab:multistft-settings}
\end{table}

\clearpage

\section{Model Details}
\label{app:model-details}

\paragraph{Generic decoders.}
Logistic regression uses a scikit-learn linear classifier on flattened aligned-channel inputs with maximum 10,000 iterations, tolerance $10^{-3}$, and random seed 42.
The MLP flattens each input sample and applies two 128-unit hidden layers with ReLU activations and 0.2 dropout.
The CNN uses three convolutional blocks with 32, 64, and 128 channels, kernel size 3, max pooling, and 0.5 dropout, followed by a 128-unit fully connected layer.
Convolutions are 1D for waveform inputs and 2D for STFT-style inputs.

We keep these model hyperparameters fixed across datasets, tasks, and the Multi-STFT and Waveform tracks.
The MLP and CNN use Adam, learning rate $10^{-4}$, batch size 200, and validation-based early stopping for up to 100 epochs.
For waveform inputs, filtering uses a 15-second context around each one-second target window to reduce boundary effects; the signal is then cropped to the target window before rereferencing, resampling, and normalization.

\paragraph{PopT-v2.}
PopT-v2 uses a PopT-style Transformer to aggregate sparse and variable electrode sets across subjects~\cite{chau_wang_2025_population}.
It has 6 layers, hidden dimension 512, 8 attention heads, 2048-dimensional feed-forward layers, multi-subject positional encoding, and a classification-token pretraining head.

We pretrain PopT-v2 from scratch on the Brain Treebank subset used by PopT: 10 subjects across 19 movie recordings~\cite{chau_wang_2025_population}.
Following TSAP's finding that PopT pretraining benefits from matching pretraining and downstream window lengths~\cite{patel2025timescale}, we use one-second Laplacian Multi-STFT windows with the duration-defined track settings and per-channel keyed standardization.
Pretraining uses a next-segment prediction and token-replacement objective, with 1-second windows sampled every 0.2 seconds, variable channel subsets of 10--100 channels, and 10\% token replacement.
Training runs for 1,000,000 steps with LAMB, learning rate $10^{-4}$, batch size 256, gradient clipping at 1.0, a 2.5\% warmup ramp-up schedule with decay factor 0.99, and bfloat16 mixed precision.
Pretraining takes 48 hours on a single A100 GPU.

For benchmark fine-tuning, we update the pretrained Transformer and a new classification head with AdamW, weight decay $10^{-2}$, learning rates $5\times10^{-5}$ and $5\times10^{-4}$ respectively, batch size 128, and 1,000 gradient-update steps.

\paragraph{HTNet.}
We evaluate an HTNet-style implementation~\cite{peterson2021htnet}, based on the EEGNet-derived architecture~\cite{lawhern2018eegnet}, on one-second waveform inputs resampled to 500 Hz.
We retain $F_1=8$, depth multiplier $D=2$, $F_2=16$, and dropout 0.5, and adapt the temporal and separable kernel lengths to 32 and 8 samples for one-second inputs.
Training uses Adam, learning rate $10^{-3}$, batch size 128, and validation-based early stopping with patience 10 for up to 100 epochs.
HTNet uses the same filtering and cropping procedure as the generic waveform decoders.

\paragraph{BaRISTA.}
BaRISTA~\cite{oganesian2025barista} is a self-supervised intracranial waveform model whose central design feature is spatial encoding and masking at configurable spatial scales beyond individual channels: temporal patch representations can be encoded and masked at the level of individual channels, parcels, or lobes, making the granularity of spatial representation an explicit modeling choice.
It is pretrained on 30 hours of Brain Treebank data, segmented into 3-second chunks sampled at 2048 Hz with 250 ms temporal patches.
Pretraining excludes all recording sessions used for testing in the benchmark.
Pretraining waveforms are Laplacian rereferenced, notch filtered, and high-pass filtered at 0.5 Hz.
Destrieux atlas labels define the spatial categories, following the original paper's best configuration: parcel-level encoding and channel-level masking.

For benchmark evaluation, we freeze the tokenizer and retain the 250 ms temporal patch size.
Fine-tuning uses learning rate $10^{-3}$, weight decay $10^{-2}$, batch size 128, and early stopping on validation ROC-AUC with patience 15 for up to 100 epochs.
We extend the Destrieux spatial encoding to Destrieux-ASEG to support more subcortical regions.

Evaluation filtering is performed separately within each recording and train, validation, or test split.
Selected windows are ordered by start time, concatenated for 0.5 Hz high-pass and notch filtering, then divided back into the original windows; the procedure does not filter the continuous recording or mix samples across splits.
BYD is filtered at 1000 Hz and upsampled to 2048 Hz; Neuroprobe and Pippi remain at 2048 Hz.
Filtering is followed by Laplacian rereferencing, robust global scaling, and per-window, per-channel temporal standardization before tokenization.

\paragraph{DIVER-1.}
We evaluate the 0.1-second-patch, Tiny-width variant of DIVER-1~\cite{han2026diver1}.
It was pretrained to reconstruct masked raw inputs and STFT/FFT features at the patch level for 32 epochs on private iEEG data and AJILE12~\cite{peterson2022ajile12}, totaling 5,310 hours.
Pretraining inputs use a 0.5 Hz high-pass filter and notch filtering.

Benchmark evaluation uses the original paper's frozen setting: learning rate $2\times10^{-3}$, weight decay $10^{-2}$, batch size 32, and early stopping on validation ROC-AUC for up to 40 epochs.
DIVER-1 uses Montreal Neurological Institute (MNI) coordinates when available; Pippi is evaluated without coordinate inputs because the benchmark release does not include Pippi MNI coordinates.

The DIVER-native evaluation route filters a 15-second context with a 0.5 Hz high-pass filter followed by notch filters at 60, 120, and 180 Hz, crops to the one-second target window, applies Laplacian rereferencing, and resamples to 500 Hz.
No additional standardization is applied.

\paragraph{Waveform preprocessing.}
Table~\ref{tab:waveform-recipes} summarizes the sampling, filtering, and normalization conventions for each system.
Baseline comparisons did not show consistent improvements from higher sampling rates.
BaRISTA and DIVER-1 retain model-specific preprocessing and pretraining exposure, so comparisons with the fixed-track baselines evaluate complete systems.

{
\begin{table}[htbp]
  \centering
  \footnotesize
  \setlength{\tabcolsep}{3pt}
  \begin{tabular}{@{}p{0.19\linewidth}p{0.09\linewidth}p{0.28\linewidth}p{0.35\linewidth}@{}}
    \toprule
    System & Input rate & Filtering procedure & Input normalization \\
    \midrule
    Logistic, MLP, CNN, HTNet & 500 Hz
      & Filter with 15-second context, then crop to the one-second target window.
      & Training-fitted robust global normalization; one median and median-absolute-deviation scale across channels and timepoints. \\
    BaRISTA & 2048 Hz
      & Concatenate selected windows in temporal order within each recording and split for filtering; upsample BYD from 1000 Hz.
      & Robust global scaling followed by per-window, per-channel temporal standardization. \\
    DIVER-1 & 500 Hz
      & Filter with 15-second context, then crop to the one-second target window.
      & No additional input standardization. \\
    \bottomrule
  \end{tabular}
  \caption{\textbf{Waveform-track preprocessing recipes.} All systems use 0.5 Hz high-pass filtering, line-noise notch filtering, and Laplacian rereferencing. Model-specific variants retain their sampling and normalization conventions.}
  \label{tab:waveform-recipes}
\end{table}
}

\subsection{Additional model coverage}
\label{app:additional-model-coverage}

MVPFormer and Brant use temporal input representations that require additional adaptation for iMINDBench's one-second naturalistic decoding windows.

\paragraph{MVPFormer.}
The released MVPFormer configuration we examined~\cite{carzaniga2026mvpformer} encodes each channel in five-second segments and combines 25 segments of temporal context.
This temporal granularity differs from iMINDBench's one-second prediction windows.
Retaining the released context would provide information beyond the benchmark window and require explicit handling of chronological fold boundaries; shortening or rescaling the input changes the signal presented to the pretrained encoder.
Including MVPFormer therefore requires a validated mapping between its temporal representation and the benchmark's prediction targets.

\paragraph{Brant.}
Brant's published preprocessing~\cite{zhang2023brant} uses six-second patches with both temporal and frequency-domain representations.
Adapting one-second recordings to this input format changes the temporal support of those representations.
In particular, stretching a short window to the expected patch length changes its effective timescale and requires care when computing frequency features.
A benchmark-compatible adaptation must establish how these representations are constructed while preserving the intended signal interpretation.

We defer inclusion of these models in the main comparison pending validation of these adaptations.
These cases motivate documenting model-specific input requirements alongside standardized benchmark protocols.

\subsection{Benchmark Runtime Estimates}

Cumulative per-job timing records are normalized to four concurrent workers.
Preprocessing and CPU fitting assume a node with two 32-core AMD EPYC 7513 processors; neural fitting assumes four NVIDIA A100 SXM4 GPUs (80 GB each), with one worker per GPU.
Estimates cover all three datasets and both folds and are not measured end-to-end wall-clock times.

\begin{table}[H]
  {
  \centering
  \small
  \begin{tabular}{@{}lrr@{}}
    \toprule
    Preprocessing route & Worker-hours & Four-worker wall time \\
    \midrule
    Multi-STFT & $\sim$5.5 h & $\sim$1.4 h \\
    Waveform & $\sim$80 h & $\sim$20 h \\
    \bottomrule
  \end{tabular}
  \caption{\textbf{Estimated preprocessing time.} Waveform preprocessing is estimated from the HTNet runs that built the cache reused by waveform Logistic, MLP, and CNN.}
  \label{tab:preprocessing-runtime}
  }
\end{table}

\begin{table}[H]
  {
  \centering
  \small
  \setlength{\tabcolsep}{5pt}
  \begin{tabular}{@{}llrr@{}}
    \toprule
    Track & Model & Fitting worker-hours & Four-worker wall time \\
    \midrule
    Multi-STFT & Logistic & 0.8 h & $\sim$0.2 h \\
    Multi-STFT & MLP & 3.9 h & $\sim$1.0 h \\
    Multi-STFT & CNN & 5.6 h & $\sim$1.4 h \\
    Multi-STFT & PopT-v2 & 72.6 h & $\sim$18.2 h \\
    Waveform & Logistic & 0.4 h & $\sim$0.1 h \\
    Waveform & MLP & 79.1 h & $\sim$19.8 h \\
    Waveform & CNN & 64.0 h & $\sim$16.0 h \\
    Waveform & HTNet & 14.1 h & $\sim$3.5 h \\
    Model-native & BaRISTA & 27.4 h & $\sim$6.8 h \\
    Model-native & DIVER-1 & 68.0 h & $\sim$17.0 h \\
    \bottomrule
  \end{tabular}
  \caption{\textbf{Estimated model-fitting time.} Fitting excludes subject loading and preprocessing.}
  \label{tab:model-fitting-runtime}
  }
\end{table}

\clearpage

\section{Leaderboard}
\label{app:leaderboard}

The interactive leaderboard is available at \url{https://imindbench.github.io/}.

\begin{figure}[H]
  \centering
  \begin{minipage}{\linewidth}
    \centering
    \includegraphics[width=0.9\linewidth]{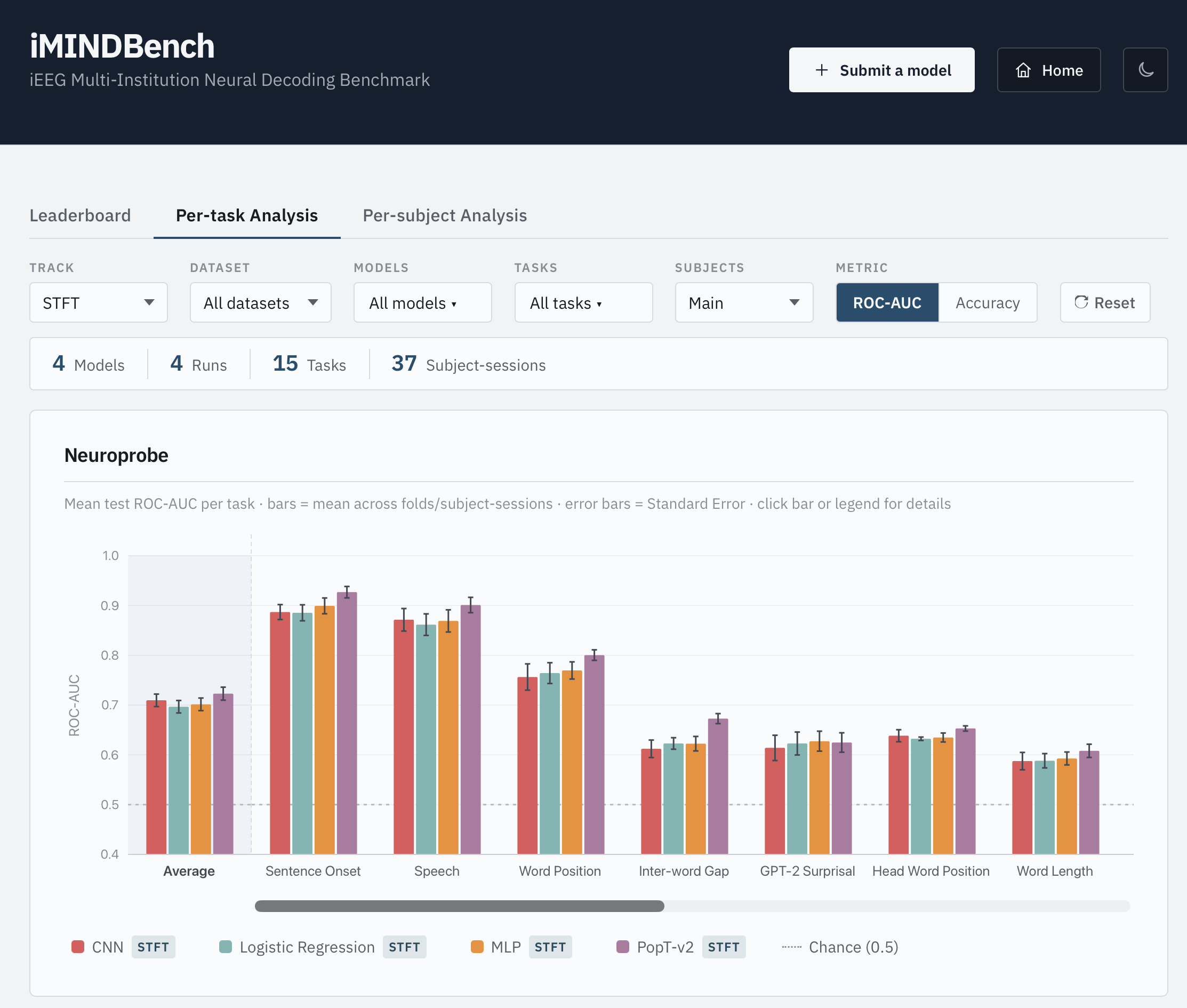}
  \end{minipage}\hfill
    \caption{\textbf{Decomposed results viewer.} The leaderboard supports comparisons across datasets, tasks, preprocessing tracks, scaling domains, and unit decodability subsets.}
    \label{fig:leaderboard}
\end{figure}

\clearpage

\section{AutoResearch Feasibility Study}
\label{app:autoresearch}

\begingroup
% Compact figure spacing keeps the study together without shrinking the body text.
\setlength{\intextsep}{6pt}

As a scoped proof of use, we applied an AutoResearch-style agentic experimentation loop~\citep{karpathy2026autoresearch} to tune the PopT-v2 pretraining recipe starting from a fixed 780k-step single-STFT checkpoint. The best selected variant improved Neuroprobe within-session decoding over the original 1M-step single-STFT baseline by $\Delta$ROC-AUC $=0.0066$ (95\% CI: 0.0021--0.0111, $p=0.0049$; Figure~\ref{fig:autoresearch}). This demonstrates how automated tuning systems may support scalable, controlled model comparisons.

\begin{figure}[H]
  \centering
  \begin{minipage}{0.58\linewidth}
    \centering
    \includegraphics[width=\linewidth]{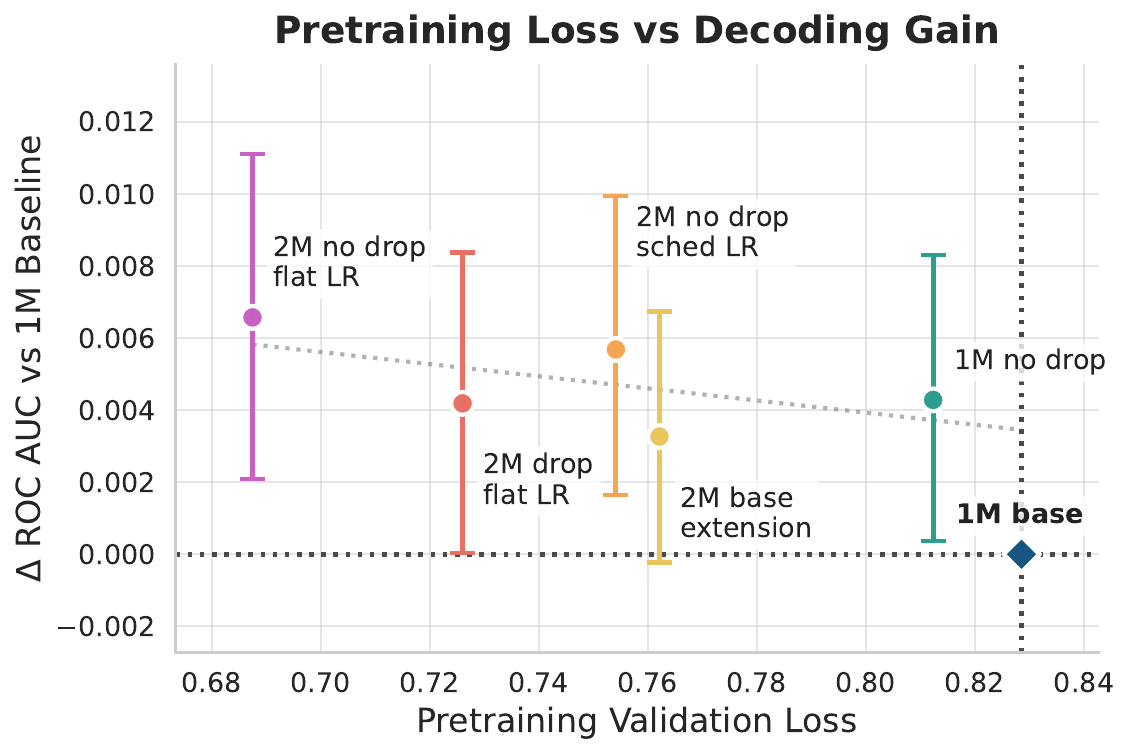}
  \end{minipage}\hfill
  \begin{minipage}{0.37\linewidth}
    \caption{\textbf{AutoResearch feasibility result.}
AutoResearch-selected PopT-v2 variants with lower pretraining validation loss (colors) generally improve Neuroprobe within-session decoding.
Points show variants evaluated across 150 paired units; error bars show 95\% bootstrap confidence intervals.}
    \label{fig:autoresearch}
  \end{minipage}
\end{figure}

AutoResearch was run in a human-in-the-loop mode from the same 780k-step single-STFT PopT-v2 checkpoint, with human decisions limited to approving phase transitions and extending promising candidates.
The search varied optimization and regularization hyperparameters (learning rate, scheduler type, batch size, gradient clipping, weight decay, and dropout) while model architecture, dataset, preprocessing track (\texttt{laplacian\_stft\_time\_pooled}), split policy (per-subject, seed 42, 1\% validation / 10\% test), and validation objective (\texttt{nsp\_replace\_only\_pretrain} loss) were held fixed.
Across 26 launched jobs (25 valid), the sweep consumed approximately 228.3 GPU-hours over approximately 184.5 active wall-clock hours.
Initial screening produced small or inconsistent effects; dropout removal, identified within the initial 780k--820k window (approximately 16.2 GPU-hours, compared to approximately 48 GPU-hours for a full 1M steps pretraining run), was the first robust improvement and was subsequently extended to 2M steps with variants of both scheduled and flat learning rates.
As shown in Figure~\ref{fig:appendix-autoresearch}, no-dropout variants separate clearly from dropout variants beyond approximately 1.2M steps, with the best model (2M no-dropout flat-LR) achieving validation loss 0.687 versus 0.829 for the original 1M baseline.
Because this model uses a single-STFT rather than the benchmark's Multi-STFT track, it is not directly comparable to Table~\ref{tab:scoreboard-all-units}; however, its within-session All-subset performance of 0.664 sits above the Multi-STFT PopT-v2 baseline (0.599 overall, 0.659 on Neuroprobe). The total compute cost was approximately \$342 (228.3 GPU-hours at \$1.50/A100-hour).

\begin{figure}[H]
  \centering
  \begin{minipage}{0.60\linewidth}
    \centering
    \includegraphics[width=\linewidth]{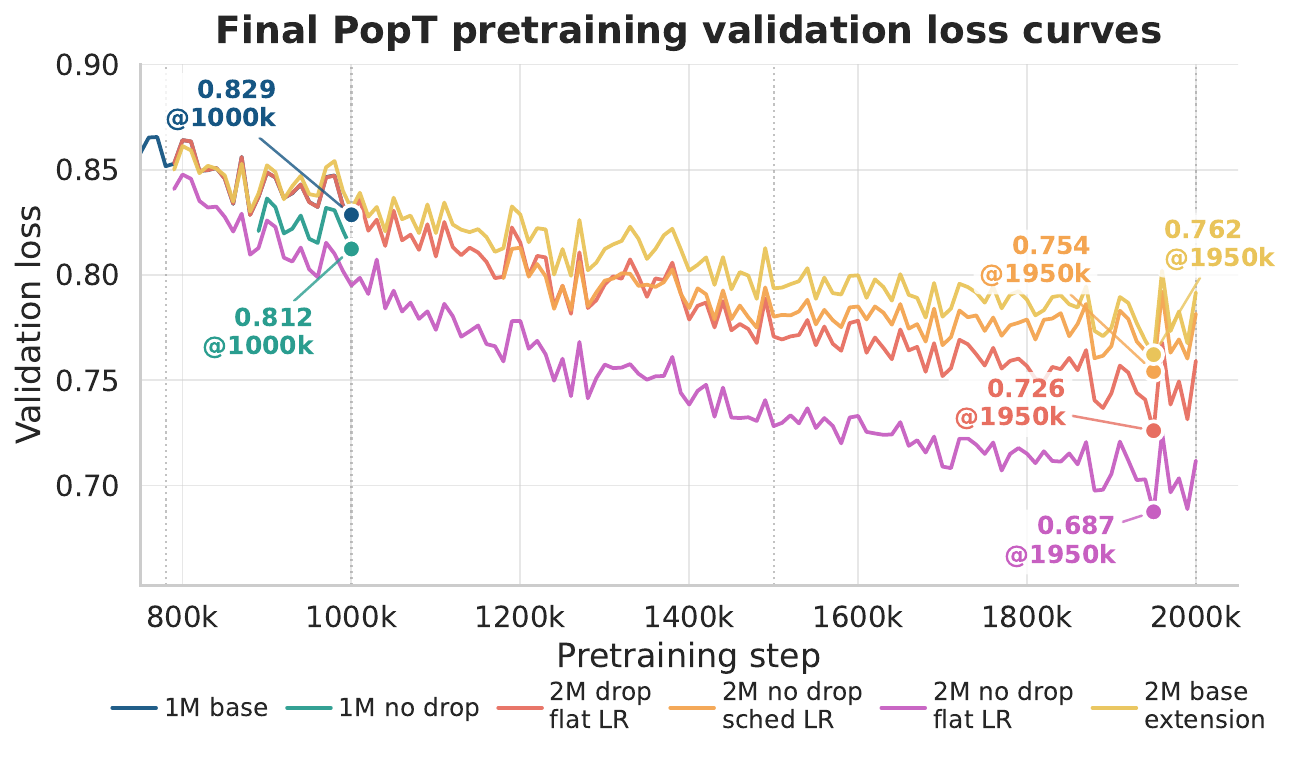}
  \end{minipage}\hfill
  \begin{minipage}{0.37\linewidth}
  \small
  \caption{\textbf{AutoResearch pretraining validation curves.}
All runs continue from a shared 780k-step checkpoint.
The 1M baseline (dropout=0.1, flat LR) and 1M no-dropout model terminate early; the four 2M continuations separate beyond approximately 1.2M steps, with no-dropout variants consistently outperforming their dropout counterparts.
The best model, 2M no-dropout flat-LR, achieves the lowest final validation loss (0.687), confirming dropout removal as the primary driver of improvement and the flat learning rate as an additional benefit at longer horizons.}
  \label{fig:appendix-autoresearch}
  \end{minipage}
\end{figure}
\endgroup

\FloatBarrier

\FloatBarrier

\end{document}